\documentclass[11pt]{article}
\usepackage{fullpage}
\usepackage{url}
\usepackage{amsmath, amssymb, amsthm, color}
\usepackage{graphicx}
\usepackage{subfigure}
\usepackage{float}
\usepackage{enumerate}
\usepackage[margin=1cm]{caption}
\newtheorem{corollary}{Corollary}

\newtheorem{lemma}{Lemma}

\newtheorem{theorem}{Theorem}

\newtheorem{claim}{Claim}
\newcommand{\cP}{\cal P}
\newcommand{\cK}{\cal K}

\begin{document}
\title{\Large Close the Gaps: A Learning-while-Doing Algorithm for a Class of Single-Product Revenue Management Problems}
\author{Zizhuo Wang\thanks{Industrial and Systems Engineering, University of Minnesota. {\tt Email:zwang@umn.edu}. Majority of this work was done while the author was a student at Stanford University} \and Shiming Deng \thanks{School of Management, Huazhong University of Science and Technology. {\tt Email:smdeng@mail.hust.edu.cn}} \and Yinyu Ye \thanks{Management Science and Engineering, Stanford University. {\tt Email:yinyu-ye@stanford.edu}}}

\maketitle \abstract{We consider a retailer selling a single product
with limited on-hand inventory over a finite selling season.
Customer demand arrives according to a Poisson process, the rate of
which is influenced by a single action taken by the retailer (such
as price adjustment, sales commission, advertisement intensity,
etc.). The relationship between the action and the demand rate is
not known in advance. However, the retailer is able to learn the
optimal action ``on the fly'' as she maximizes her total expected
revenue based on the observed demand reactions.

Using the pricing problem as an example, we propose a dynamic
``learning-while-doing'' algorithm that only involves function value
estimation to achieve a near-optimal performance. Our algorithm
employs a series of shrinking price intervals and iteratively tests
prices within that interval using a set of carefully chosen
parameters. We prove that the convergence rate of our algorithm is
among the fastest of all possible algorithms in terms of asymptotic
``regret'' (the relative loss comparing to the full information
optimal solution). Our result closes the performance gaps between
parametric and non-parametric learning and between a post-price
mechanism and a customer-bidding mechanism. Important managerial
insight from this research is that the values of information on both
the parametric form of the demand function as well as each
customer's exact reservation price are less important than prior
literature suggests. Our results also suggest that firms would be
better off to perform dynamic learning and action concurrently
rather than sequentially.}

\section{Introduction}
\label{section:introduction}

Revenue management is a central problem for many industries such as
airlines, hotels, and retailers. In revenue management problems, the
availability of products is often limited in quantity and/or time,
and customer demand is either unknown or uncertain. However, demands
can be influenced by actions such as price adjustment, advertisement
intensity, sales person compensation, etc. Retailers are interested
in methods to select an optimal action plan that maximizes their
revenue in such environments.

Most existing research in revenue management assumes that the
functional relationship between the demand distribution (or the
instantaneous demand rate) and retailers' actions is known to the
decision makers. This relationship is then exploited to derive
optimal policies. However, in reality, decision makers seldom
possess such information. This is especially true when a new product
or service is provided at a new location or the market environment
has changed. In light of this, some recent research has proposed
methods that allow decision makers to probe the demand functions
while optimizing their policies based on learning.

There are two types of learning models: parametric and
non-parametric.  In parametric learning, prior information has been
obtained about which parametric family the demand function belongs
to. In this case, decision makers take actions ``on the fly'' while
updating their estimations about the underlying parameters. On the
other hand, non-parametric learning assumes no properties of the
demand function except some basic regularity conditions, and it is
the decision maker's task to learn the demand curve with such
limited information. Intuitively, non-parametric approaches are more
complex than the parametric ones because the non-parametric function
space is much larger. However, the exact performance difference
between these two models is not clear and several questions are to
be answered: First, what are the ``optimal'' learning strategies for
each setting? Second, what are the minimal revenue losses that will
be incurred over all possible strategies? Third, how valuable is the
information that the demand function belongs to a particular
parametric family? Moreover, it seems quite advantageous for the
sellers to know each customer's exact valuation rather than only
observing ``yes-or-no'' purchase decisions. But exactly how much
benefit is it?

In this paper, we provide answers to these questions using an
example in which the retailer's sole possible action is to control
the price. 
In such a problem, a retailer sells a limited inventory over a
finite sales season. The demand is a Poisson process whose intensity
is controlled by the prevailing price posted by the retailer. The
relationship between the price and the demand rate is not known to
the retailer and she can only learn this information through
observing realized demand at the price she sets. Specifically, we
are interested in a non-parametric setting in which the retailer
knows little about the demand function except some regularity
conditions. The objective of the retailer is to maximize her revenue
over the entire selling season.


As discussed in prior literature, the key of a good pricing
algorithm under demand uncertainty lies in its ability to balance
the tension between demand learning (exploration) and near-optimal
pricing (exploitation). The more time one spends in price
exploration, the less time remains to exploit the knowledge to
obtain the optimal revenue. On the other hand, if exploration is not
done sufficiently, then one may not be able to find a price good
enough to achieve a satisfactory revenue. This is especially true in
the non-parametric setting where it is harder to infer structural
information from the observed demand. In prior literature,
researchers have proposed price learning algorithms with separated
learning and doing phases, in which a grid of prices are tested and
then the optimal one is used for pricing. Theoretical results are
established to show that those algorithms achieve asymptotic
optimality at a decent rate, see Besbes and Zeevi \cite{besbes}.

The main contribution of our paper is to propose a dynamic
price-learning algorithm that iteratively performs price
experimentation within a shrinking series of intervals that always
contain the optimal price (with high probability). We show that our
dynamic price-learning algorithm has a tighter upper bound on the
asymptotic regret (the relative loss to the clairvoyant optimal
revenue) than the one-time grid learning strategy. By further
providing a worst-case example, we prove that our algorithm provides
the near best asymptotic performance over all possible pricing
policies. Our result provides important managerial insights to
revenue management practitioners: we should not separate price
experimentation from exploitation, instead, we should combine
``learning'' and ``doing'' in a concurrent procedure to achieve the
smallest revenue loss.

We further summarize our contributions in the following:
\begin{enumerate}

\item Under some mild regularity
conditions on the demand function, we show that our pricing policy
achieves a regret of $O^*(n^{-1/2})$\footnote{We use the notation
$f(n)=O^*(g(n))$ to mean that for some constant $C_1$ and $C_2$,
$f(n)\le C_lg(n)(\log{n})^{C_2}$.}, uniformly across all demand
functions. This result improves the best-known bound (of the
asymptotic regret) by Besbes and Zeevi \cite{besbes} for both the
non-parametric learning (the best known bound was
$O^*\left(n^{-1/4}\right)$) and the parametric learning (the best
known bound was $O^*\left(n^{-1/3}\right)$) in this context. One
consequence of our result is that it closes the efficiency gap
between parametric and non-parametric learning, implying that the
value of knowing the parametric form of the demand function in this
problem is marginal when the best algorithm is adopted. In this
sense, our algorithm could save firms' efforts in searching for the
right parametric form of the demand functions.

\item Our result also closes the gap between two price revealing mechanisms in revenue management: the customer-bidding mechanism and the
post-price mechanism. Agrawal et al \cite{shipra} obtained a
near-optimal dynamic learning algorithm with $O^*(n^{-1/2})$ regret
under the former mechanism (in a slightly different setting).
However, in post-price models, the previous best algorithm by Besbes
and Zeevi \cite{besbes} achieves a regret of $O^*(n^{-1/4})$. It was
unclear how large is the gap between the two pricing mechanisms.
Surprisingly, our algorithm shows that although the post-price
mechanism extracts much less information from each individual
customer's valuation of the product, it can achieve the same order
of asymptotic performance as that of customer-bidding mechanisms.
Therefore, our result reassures the usage of the post-price
mechanism, which is much more widespread in practice.

\item On the methodology side, our algorithm provides a new form of dynamic learning
mechanism. In particular, we do not separate the ``learning'' and
``doing'' phases; instead, we integrate ``learning'' and ``doing''
together by considering a shrinking series of price intervals. This
concurrent dynamic is the key to achieve a perfect balance between
price exploration and exploitation, and thus achieve the near
maximum efficiency. We believe that this method may be applied to
problems with even more complex structures.
\end{enumerate}
%

\section{Literature Review}
\label{section:literaturereview}

Pricing strategies have been an important research area in revenue
management and there is abundant literature on this topic. We refer
to Bitran and Caldentey \cite{bitran}, Elmaghraby and Keskinocak
\cite{elmaghraby} and Talluri and van Ryzin \cite{talluribook} for a
comprehensive review on this subject. Prior research has mostly
focused on the cases where the functional relationship between the
price and demand (also known as the demand function) is given to the
decision maker. For example, Gallego and van Ryzin \cite{gallego1}
presented a foundational work in such a setting where the structure
of the demand function is exploited and the optimal pricing policies
are analyzed.

Although knowing the exact demand function is convenient for
analysis, decision makers in practice do not usually have such
information. Therefore, much recent literature addresses the dynamic
pricing problems with unknown demand function. The majority of these
work take the parametric approach, e.g., Lobo and Boyd \cite{lobo},
Bertsimas and Parekis \cite{bertsimas}, Araman and Caldentey
\cite{araman}, Aviv and Pazgal \cite{aviv}, Carvalho and Puterman
\cite{carvalho}, Farias and Van Roy \cite{farias}, Broder and
Rusmevichientong \cite{broder}, den Boer and Zwart
\cite{denboer_finite_inventory, denboer1} and Harrison et al
\cite{harrison}. In these pieces of work, the demand function is
assumed to take a certain parametric form, and the design of
algorithm is revolved around learning and estimating the underlying
parameters. Although such an approach simplifies the problem to some
degree, the restriction to a certain demand function family may
incur model misspecification risk. As shown in Besbes and Zeevi
\cite{besbes}, misspecifying the demand family may lead to revenue
that is far from the optimal. In such cases, a non-parametric
approach would be preferred because it does not commit to a single
family of demand functions.

The main difficulty facing the non-parametric approach is its
tractability and efficiency. And most research revolves around this
question. Several studies consider a model that the customers are
chosen adversarially, e.g. Ball and Queyranne \cite{ball} and
Perakis and Roels \cite{perakis}. However, their models take a
relatively conservative approach and no learning is involved. In
another paper by Lim and Shanthikumar \cite{lim}, they consider
dynamic pricing strategies that are robust to an adversarial at
every point in the stochastic process. Again, this approach is quite
conservative and the main theme is about robustness rather than
demand learning.

The work that is closest to our paper is that of Besbes and Zeevi
\cite{besbes} where the authors consider demand learning in both
parametric and non-parametric cases. They propose learning
algorithms for both cases and show that there is a gap in
performance between them. They also provide lower bounds for the
revenue loss in both cases. In this paper, we improve their bounds
for both cases and close the gap between them. In particular, they
consider algorithms that separate learning and doing phases where
price experimentation is performed exclusively during the learning
phase (except the parametric case with a single parameter). In our
paper, learning and doing are dynamically integrated: we keep
shrinking a price interval that contains the optimal price and keep
learning until we guarantee that the revenue achieved by applying
the current price is near-optimal. Although our model resembles
theirs, our algorithm is quite different and the results are
stronger.

More recently, Besbes and Zeevi \cite{besbes_linear} propose a new
dynamic pricing algorithm, in which the seller pretends that the
underlying demand function is linear and chooses price to maximize a
proxy revenue function based on the estimated linear demand
function. They show that although the model might be misspecified,
one can still achieve a regret of $O^*(n^{-1/2})$. Their result is
novel and important. However, they require a slightly stricter
condition on the true underlying demand function than ours.
Moreover, their model doesn't allow inventory constraints and only
considers a discrete time framework. As pointed out in
\cite{denboer_finite_inventory} and shown in our paper, the presence
of inventory constraints and the continuous time framework could
bring significant differences to the analysis and performance of
pricing algorithms. Similar differences also exist between our work
and that of Kleinberg and Leighton \cite{kleinberg}.

Other related literature that focus on the exploration-exploitation
trade-off in sequential optimization under uncertainty are from the
study of the multi-armed bandit problem: see Lai and Robbins
\cite{lai}, Agrawal \cite{agrawalbandit}, Auer et al \cite{auer} and
references therein. Although our study shares similarity in ideas
with the multi-armed bandit problem, the problem we consider has a
continuous learning horizon (the time is continuous), continuous
learning space (the possible demand function is continuous) and
continuous action space (the price set is continuous). These
features with the presence of inventory constraint distinguish our
algorithm and analysis from theirs.

\section{Problem Formulation}
\label{section:model}

\subsection{Model and Assumptions}
In this paper, we consider the problem of a monopolist selling a
single product in a finite selling season $T$. The seller has a
fixed inventory $x$ at the beginning and no recourse actions on the
inventory can be made during the selling season. During the selling
season, customers arrive according to a Poisson process with an
instantaneous demand rate $\lambda_t$ at any time $t$. In our model,
we assume that $\lambda_t$ is solely determined by the price offered
at time $t$, that is, we can write $\lambda_t=\lambda(p(t))$ where
$p(t)$ is the price at time $t$. At time $T$, the sales will be
terminated and there is no salvage value for the remaining items (as
shown in \cite{gallego1}, the zero salvage value assumption is
without loss of generality).

We assume the set of feasible prices is an interval
$[\underline{p},\overline{p}]$ with an additional cut-off price
$p_\infty$ such that $\lambda(p_\infty)=0$. The demand rate function
$\lambda(p)$ is assumed to be decreasing in $p$ and has an inverse
function $p=\gamma(\lambda)$. The revenue rate function $r(\lambda)
= \lambda\gamma(\lambda)$ is assumed to be concave in $\lambda$.
These assumptions are quite standard and such demand functions are
called the ``regular'' demand functions in the revenue management
literature \cite{gallego1}.

In addition, we make the following assumptions
on the demand rate function $\lambda(p)$ and the revenue rate function $r(\lambda)$:\\

{\noindent\bf Assumption 1.}
 For some positive constants $M$, $K$, $m_L$ and $m_U$,
\begin{enumerate}
\item Boundedness: $|\lambda(p)|\le M$ for
all $p\in[\underline{p},\overline{p}]$;
\item Lipschitz continuity: $\lambda(p)$
and $r(\lambda(p))$ are Lipschitz continuous with respect to $p$
with a factor $K$. Also, the inverse demand function
$p=\gamma(\lambda)$ is Lipschitz continuous in $\lambda$ with a factor
$K$;
\item Strict concavity and differentiability: $r''(\lambda)$ exists and $-m_L\le r''(\lambda)\le -m_U<0$
for all $\lambda$ in the range of $\lambda(p)$ for
$p\in[\underline{p},\overline{p}]$.
\end{enumerate}
In the following, let $\Gamma = \Gamma (M,K,m_L,m_U)$ denote the set
of demand functions satisfying the above assumptions with the
corresponding coefficients. Assumption 1 is quite mild and has been
adopted in several prior revenue management literature, see, e.g.,
\cite{broder,kleinberg}. It holds for many commonly-used demand
function classes including linear, exponential, and logit demand
functions.

In our model, the seller does not know the true demand function
$\lambda$, the only knowledge she has is that it belongs to
$\Gamma$. Note that $\Gamma$ doesn't need to have any parametric
representation. Therefore, our model is robust in terms of the
choice of the demand function family.

\subsection{Minimax Regret Objective}
To evaluate the performance of any pricing algorithm, we adopt the
minimax regret objective formalized in \cite{besbes}. We call a
pricing policy $\pi = (p(t): 0\le t\le T)$ admissible if it is a
non-anticipating price process that is defined on
$[\underline{p},\overline{p}]\cup \{p_\infty\}$, and satisfies the
inventory constraint, that is
\begin{eqnarray*}
\int_0^TdN^\pi(s)\le x\quad\quad\mbox{with probability }1,
\end{eqnarray*}
where $N^\pi(t)=N\left(\int_{0}^t\lambda(p(s))ds\right)$ denotes the
cumulative demand up to time $t$ using policy $\pi$.

We denote the set of admissible pricing policies by $\cP$.  The
expected revenue generated by a policy $\pi$ is defined by
\begin{equation}\label{expectedrevenue}
J^\pi(x,T;\lambda) = E\left[\int_{0}^Tp(s)dN^\pi(s)\right].
\end{equation}
Here, the presence of $\lambda$ in $J^\pi(x,T;\lambda)$ means that
the expectation is taken with respect to the demand function
$\lambda$. Given a demand function $\lambda$, we wish to find the
optimal admissible policy $\pi^*$ that maximizes
(\ref{expectedrevenue}). In our model, since we don't know the exact
$\lambda$, we seek $\pi\in\cP$ that performs as close to $\pi^*$ as
possible.

However, even if the demand function $\lambda$ is known, computing
the expected value of the optimal policy is hard. It involves
solving a Bellman equation resulting from a dynamic program.
Fortunately, as shown in \cite{besbes, gallego1}, we can obtain an
upper bound for the expected value for any policy via considering a
full-information deterministic optimization problem. Define:
\begin{equation}
\begin{array}{llll}\label{deterministic}
J^D(x,T;\lambda ) = & \sup         & \int_0^Tr(\lambda(p(s))) ds      & \\
                    & \mbox{s.t.}  & \int_0^T \lambda(p(s)) ds \le x  & \\
      &       & p(s)\in[\underline{p},\overline{p}]\cup \{p_\infty\} &
             \forall s\in [0,T].
\end{array}
\end{equation}
In (\ref{deterministic}) all the stochastic processes are
substituted by their means. In \cite{besbes}, the authors showed
that $J^D(x,T;\lambda)$ provides an upper bound on the expected
revenue generated by any admissible pricing policy $\pi$, that is,
$J^\pi(x,T;\lambda)\le J^D(x,T;\lambda)$, for all $\lambda\in\Gamma$
and $\pi\in\cP$. With this relaxation, we can define the regret
$R^\pi(x,T;\lambda)$ for any given demand function
$\lambda\in\Gamma$ and policy $\pi\in\cP$ by
\begin{equation}\label{regret}
R^\pi(x,T;\lambda) = 1 -
\frac{J^\pi(x,T;\lambda)}{J^D(x,T;\lambda)}.
\end{equation}
Clearly, $R^\pi(x,T;\lambda)$ is always greater than $0$. And by
definition, the smaller the regret, the closer $\pi$ is to the
optimal policy. However, since the decision maker does not know the
true demand function, it is attractive to obtain a pricing policy
$\pi$ that achieves small regrets across all the underlying demand
function $\lambda\in\Gamma$. To capture this, we consider the
``worst-case'' regret. That is, the decision maker chooses a pricing
policy $\pi$, and the nature picks the worst possible demand
function for that policy:
\begin{eqnarray*}
\sup_{\lambda\in\Gamma}R^\pi(x,T;\lambda).
\end{eqnarray*}
Our goal is to minimize the worst-case regret:
\begin{equation}\label{minimizeworstcaseregret}
\inf_{\pi\in \cP}\sup_{\lambda\in\Gamma}R^\pi(x,T;\lambda).
\end{equation}
Unfortunately, it is hard to evaluate
(\ref{minimizeworstcaseregret}) for any finite size problem. In this
work, we adopt the widely-used asymptotic performance analysis. We
consider a regime in which both the size of the initial inventory,
as well as the potential demand, grow proportionally large. In a
problem with size $n$, the initial inventory and the demand function
are given by\footnote{It is worth pointing out that one could also
define the size $n$ problem by fixing $\lambda(\cdot)$ while
enlarging $x$ and $T$ by a factor of $n$. Our proposed algorithm
still works in that case, with the length of each learning period to
be $n$ times the length of the learning period in the current
algorithm. We choose the current way of presentation to maintain
consistency with the literature.}:
\begin{eqnarray*}
x_n = nx \mbox{ and } \lambda_n(\cdot) = n \lambda(\cdot).
\end{eqnarray*}
Define $J^D_n(x,T;\lambda) = J^D(nx, T, n\lambda) = nJ^D(x, T,
\lambda)$ to be the deterministic optimal solution for the problem
with size $n$ and $J^\pi_n(x,T;\lambda) = J^\pi(nx, T, n\lambda)$ to
be the expected value of a pricing policy $\pi$ when it is applied
to a problem with size $n$. The regret for the size-$n$ problem
$R^\pi_n(x,T;\lambda)$ is therefore defined by
\begin{equation}\label{regretn}
R^\pi_n(x,T;\lambda) = 1 -
\frac{J^\pi_n(x,T;\lambda)}{J^D_n(x,T;\lambda)},
\end{equation}
and our objective is to study the asymptotic behavior of
$R^\pi_n(x,T;\lambda)$ as $n$ grows large and design an algorithm
with small asymptotic regret.

\section{Main Results: Dynamic Pricing Algorithm}
\label{section:mainresult}

In this section, we introduce our dynamic pricing algorithm. Before
we state our main results, it is useful to discuss some basic
structural intuitions of this problem.
\subsection{Preliminary Ideas}
\label{main result:structural insights} Consider the
full-information deterministic problem (\ref{deterministic}). As
shown in \cite{besbes}, the optimal solution to
(\ref{deterministic}) is given by
\begin{equation}\label{optimalpricesolution}
p(t) = p^D = \max\{p^u,p^c\}
\end{equation} where
\begin{equation}\label{pu}
p^u = arg\max_{p\in[\underline{p},\overline{p}]}\{r(\lambda(p))\},
\end{equation}
\begin{equation}\label{pc}
p^c =
arg\min_{p\in[\underline{p},\overline{p}]}|\lambda(p)-\frac{x}{T}|.
\end{equation}
Here, the superscript $u$ stands for ``unconstrained'' and
superscript $c$ stands for ``constrained''. As shown in
(\ref{optimalpricesolution}), the optimal price is either the
revenue maximizing price, or the inventory depleting price,
whichever is larger. The following important lemma is proved in
\cite{gallego1}.
\begin{lemma}\label{lemma:gallego1}
Let $p^D$ be the optimal deterministic price when the underlying
demand function is $\lambda$. Let $\pi^D$ to be the pricing policy
that uses the deterministic optimal price $p^D$ throughout the
selling season until there is no inventory left. Then
$R_n^{\pi^D}(x,T,\lambda) = O(n^{-1/2})$.
\end{lemma}
Lemma \ref{lemma:gallego1} says that if one knows $p^D$, then simply
applying this price can achieve asymptotically optimal performance.
Therefore, the idea of our algorithm is to find an estimate $p^D$
close enough to the true one efficiently, using empirical
observations on hand. In particular, under Assumption 1, we know
that if $p^D = p^u
> p^c$, then
\begin{eqnarray}\label{localpu}
|r(p)-r(p^D)|\le \frac{1}{2}M_L(p-p^D)^2
\end{eqnarray}
for $p$ close to $p^D$ while if $p^D = p^c \ge p^u$, then
\begin{eqnarray}\label{localpc}
|r(p)-r(p^D)| \le K|p-p^D|
\end{eqnarray}
for $p$ close to $p^D$. In the following discussion, without loss of
generality, we assume $p^D\in (\underline{p}, \overline{p})$. Note
that this can always be achieved by choosing a large interval of
$[\underline{p}, \overline{p}]$.

\subsection{Dynamic Pricing Algorithm}
\label{main results: theorems} We first state our main results as
follows:
\begin{theorem}\label{th:maintheorem}
Let Assumption 1 hold for $\Gamma = \Gamma(M,K,m_L,m_U)$. Then there
exists an admissible policy $\pi$ generated by Algorithm DPA, such
that for all $n\ge 1$,
\begin{eqnarray*}
\sup_{\lambda\in \Gamma}R_n^{\pi}(x,T;\lambda) \le
\frac{C(\log{n})^{4.5}}{\sqrt{n}}
\end{eqnarray*}
for some constant $C$.
\end{theorem}
Here $C$ only depends on $M, K, m_L, m_U$, the initial inventory $x$
and the length of time horizon $T$. The exact dependence is quite
complex and thus omitted. A corollary of Theorem
\ref{th:maintheorem} follows from the relationship between the
non-parametric model and the parametric one:
\begin{corollary}\label{cor:maintheorem}
Assume $\Gamma$ is a parameterized demand function family satisfying
Assumption 1. Then there exists an admissible policy $\pi$ generated
by Algorithm DPA, such that for all $n\ge 1$,
\begin{eqnarray*}
\sup_{\lambda\in \Gamma}R_n^{\pi}(x,T;\lambda) \le
\frac{C(\log{n})^{4.5}}{\sqrt{n}}
\end{eqnarray*}
for some constant $C$ that only depends on the coefficients in
$\Gamma$, $x$ and $T$.
\end{corollary}

Now we describe our dynamic pricing algorithm. As alluded in Section
\ref{main result:structural insights}, we aim to learn $p^D$ through
price experimentations. Specifically, our algorithm will be able to
distinguish whether ``$p^u$'' or ``$p^c$'' is optimal. Meanwhile we
keep a shrinking interval containing the optimal price with high
probability until a certain accuracy is achieved.

\hrulefill\\

{\bf\noindent Algorithm DPA (Dynamic Pricing Algorithm) :\\}

{\noindent\bf Step 1. Initialization:}

\begin{enumerate}[(a)]
\item Consider a sequence of $\tau_i^u, \kappa_i^u$, $i =
    1,2,...,N^u$ and $\tau_i^c, \kappa_i^c$, $i=1,2,...,N^c$
    ($\tau$ and $\kappa$ represent the length of each learning period
    and the number of different prices to be tested in each learning period, respectively.
    Their values along with the values of $N^u$ and $N^c$ are defined in (\ref{accuratetauu}) - (\ref{differencepc}), (\ref{conditionforNu}) and (\ref{conditionforNc})).
    Define $\underline{p}_1^u  = \underline{p}_1^c = \underline{p}$
    and
    $\overline{p}_1^u = \overline{p}_1^c =\overline{p}$. Define $t_i^u = \sum_{j=1}^i\tau_j^u$, for $i = 0$ to $N^u$ and $t_i^c = \sum_{j=1}^{i}\tau_j^c$,
    for $i=0$ to $N^c$;
\end{enumerate}
{\noindent\bf Step 2. Learn $p^u$ or determine $p^c > p^u$:}\\
{\noindent For $i = 1$ to $N^u$ do}
\begin{enumerate}[(a)]
\item Divide $[\underline{p}_i^u,\overline{p}_i^u]$ into $\kappa_i^u$
equally spaced intervals and let $\{p_{i,j}^u, j =
1,2,...,\kappa_i^u\}$ be the left endpoints of these intervals;
\item Divide the time interval $[t_{i-1}^u,t_i^u]$ into $\kappa_i^u$ equal
parts and define $$\Delta_i^u = \frac{\tau_i^u}{\kappa_i^u},
\quad\quad t_{i,j}^u = t_{i-1}^u + j \Delta_i^u, \quad\quad j =
0,1,...,\kappa_i^u;$$
\item For $j$ from 1 to $\kappa_i^u$, apply $p_{i,j}^u$ from time $t_{i,j-1}^u$ to $ t_{i,j}^u$. If inventory runs
out, then apply $p_{\infty}$ until time $T$ and STOP;
\item Compute $$\hat{d}(p_{i,j}^u) = \frac{\mbox{total demand over  } [t_{i,j-1}^u,
        t_{i,j}^u)}{\Delta_i^u} ,\quad j = 1,...,\kappa_i^u;$$
\item Compute
\begin{eqnarray}\label{pctarget1}
\hat{p}^u_i = arg\max_{1\le j \le \kappa_i^u}\{p_{i,j}^u\hat{d}
(p_{i,j}^u)\}\quad\mbox{   and    }\quad\hat{p}^c_i = arg\min_{1\le
j\le\kappa_i^u}|\hat{d}(p_{i,j}^u)-x/T|;
\end{eqnarray}
\item If
\begin{equation}\label{distinguish}
\hat{p}_i^c > \hat{p}_i^u +
2\sqrt{\log{n}}\cdot\frac{\overline{p}_i^u-\underline{p}_i^u}{\kappa_i^u}
\end{equation}
then break from Step 2, enter Step 3 and set $i_0 = i$;\\
Otherwise, set $\hat{p}_i = \max\{\hat{p}^c_i,\hat{p}^u_i\}$. Define
\begin{equation}\label{nextlowerboundu}
\underline{p}_{i+1}^u = \hat{p}_i - \frac{\log{n}}{3} \cdot
\frac{{\overline{p}_i^u}-{\underline{p}_i^u}}{\kappa_i^u}
\end{equation} and
\begin{equation}\label{nextupperboundu}
\overline{p}_{i+1}^u  = \hat{p}_i + \frac{2\log{n}}{3} \cdot
\frac{{\overline{p}_i^u}-{\underline{p}_i^u}}{\kappa_i^u}.
\end{equation}
And define the price range for the next iteration
$$I_{i+1}^u = [\underline{p}_{i+1}^u,\overline{p}_{i+1}^u].$$
Here we truncate the interval if it doesn't lie inside the feasible
set $[\underline{p},\overline{p}]$;
\item If $i=N^u$, then enter Step 4(a);
\end{enumerate}

{\noindent\bf Step 3. Learn $p^c$ when $p^c > p^u$:}\\
{\noindent For $i = 1$ to $N^c$ do}
\begin{enumerate} [(a)]
\item Divide $[\underline{p}_i^c,\overline{p}_i^c]$ into $\kappa_i^c$
equally spaced intervals and let $\{p_{i,j}^{c}, j =
1,2,...,\kappa_i^c\}$ be the left endpoints of these intervals;
\item Define $$\Delta_i^c = \frac{\tau_i^c}{\kappa_i^c}, \quad\quad t_{i,j}^c =
t_{i-1}^c + j\Delta_i^c + t_{i_0}^u, \quad\quad j =
0,1,...,\kappa_i^c;$$
\item For $j$ from $1$ to $\kappa_i^c$, apply $p_{i,j}^c$ from time $t_{i,j-1}^c$ to $ t_{i,j}^c$. If inventory runs
out, then apply $p_{\infty}$ until time $T$ and STOP;
\item Compute $$\hat{d}(p_{i,j}^c) = \frac{\mbox{total demand over  } [t_{i,j-1}^c,
t_{i,j}^c)}{\Delta_i^c} ,\quad j = 1,...,\kappa_i^c;$$
\item Compute
\begin{equation}\label{pctarget}
\hat{q}_i = arg\min_{1\le j\le\kappa_i^c} |\hat{d}(p_{i,j}^c)-x/T|.
\end{equation}
Define
\begin{equation}\label{nextlowerboundc}
\underline{p}_{i+1}^c = \hat{q}_i - \frac{\log{n}}{2} \cdot
\frac{{\overline{p}_i^c}-{\underline{p}_i^c}}{\kappa_i^c}
\end{equation} and
\begin{equation}\label{nextupperboundc}
\overline{p}_{i+1}^c = \hat{q}_i + \frac{\log{n}}{2} \cdot
\frac{{\overline{p}_i^c}-{\underline{p}_i^c}}{\kappa_i^c}.
\end{equation}
And define the price range for the next iteration
$$I_{i+1}^c = [\underline{p}_{i+1}^c,\overline{p}_{i+1}^c].$$
Here we truncate the interval if it doesn't lie inside the feasible
set of $[\underline{p},\overline{p}]$;
\item If $i = N^c$, then enter Step 4(b);
\end{enumerate}
{\bf\noindent Step 4. Apply the learned price:}
\begin{enumerate}[(a)]
\item Define $\tilde{p} = \hat{p}_{N^u} + 2\sqrt{\log{n}} \cdot
\frac{{\overline{p}_{N^u}^u}-{\underline{p}_{N^u}^u}}{\kappa_{N^u}^u}$.
Use $\tilde{p}$ for the rest of the selling season until the
inventory runs out;
\item Define $\tilde{q} = \hat{q}_{N^c}$. Use $\tilde{q}$ for the rest
of the selling season until the inventory runs out.
\end{enumerate}
\hrulefill

Now we explain the idea behind our algorithm before we proceed to
proofs. In this algorithm, we divide the time interval into a
carefully selected number of pieces. In each piece, we test a grid
of prices on a certain price interval. Based on the empirical
observations, we shrink the price interval to a smaller subinterval
that still contains the optimal price (with high probability), and
enter the next time interval with the smaller price range. We repeat
the shrinking procedure until the price interval is small enough so
that the desired accuracy is achieved.

Recall that the optimal deterministic price $p^D$ is equal to the
maximum of $p^u$ and $p^c$, where $p^u$ and $p^c$ are solved from
(\ref{pu}) and (\ref{pc}) respectively. As shown in (\ref{localpu})
and (\ref{localpc}), the local behavior of the revenue rate function
is quite different around $p^u$ and $p^c$: the former one resembles
a quadratic function while the latter one resembles a linear
function. This difference requires us to use different shrinking
strategies for the cases when $p^u > p^c$ and $p^c > p^u$. This is
why we have two learning steps (Step 2 and 3) in our algorithm.
Specifically, in Step 2, the algorithm works by shrinking the price
interval until either a transition condition (\ref{distinguish}) is
triggered or the learning phase is terminated. We show that (in
Lemma \ref{lemma:containsprice} in Section \ref{section:proof}),
when the transition condition (\ref{distinguish}) is triggered, with
high probability, the optimal solution to the deterministic problem
is $p^c$. Otherwise, if we terminate learning before the condition
is triggered, we know that $p^u$ is either the optimal solution to
the deterministic problem or it is close enough so that using $p^u$
will also yield a near-optimal revenue. When (\ref{distinguish}) is
triggered, we switch to Step 3, in which we use a new set of
shrinking and price testing parameters. Note that in Step 3, we
start from the initial price interval rather than the current
interval obtained. This is not necessary but for the ease of
analysis. Both Step 2 and Step 3 (if it is invoked) must terminate
in a finite number of iterations (we prove this in Lemma
\ref{lemma:boundedstep}).

In the end of the algorithm, a fixed price is used for the remaining
selling season (Step 4) until the inventory runs out. In fact,
instead of applying a fixed price in step 4, one may continue
learning using our shrinking strategy. However, it will not further
improve the asymptotic performance of our algorithm. Also in the
algorithm, we always use $x/T$ as the target when learning $p^c$ (in
(\ref{pctarget1}) and (\ref{pctarget})), but not the updated
remaining inventory and time. The reason is explained by Lemma
\ref{lemma:gallego1}, which states that learning $p^D$ (thus using
$x/T$) is sufficient to get an optimal performance. This in fact
leads to an interesting observation, that is, capturing the
fluctuation of the demand is a relative secondary consideration in
pricing in an uncertain demand environment.

Another thing to note is that the ``next'' intervals defined in
(\ref{nextlowerboundu}) and (\ref{nextupperboundu}) in our algorithm
are not symmetric about the current best estimate. Similarly in Step
4(a), we use an adjusted price for the remaining selling season.
These adjustments are to make sure that the inventory consumption
can be adequately upper bounded. Meanwhile the adjustments are kept
small enough so that the revenue is maintained. The detailed
reasoning of these adjustments will be illustrated in Section
\ref{section:proof}.

In the following, we define $\tau_i^u, \kappa_i^u, N^u, \tau_i^c,
\kappa_i^c$ and $N^c$. Without loss of generality, we assume $T=1$
and $\overline{p}-\underline{p} = 1$ in the following discussion. We
first provide a set of relationships we want $(\tau_i^u,\kappa_i^u)$
and $(\tau_i^c, \kappa_i^c)$ to satisfy. Then we explain the meaning
of each relationship and derive a set of parameters that satisfy
these relationships. We use the notation $f\sim g$ to mean that $f$
and $g$ are of the same order in $n$.

The set of relationships that we want
$(\tau_i^u,\kappa_i^u)_{i=1}^{N^u}$ to satisfy are:
\begin{equation}\label{condition1u}
\left(\frac{\overline{p}_i^u-\underline{p}_i^u}{\kappa_i^u}\right)^2\sim\sqrt{\frac{\kappa_i^u}{n\tau_i^u}},\quad\forall
i = 2,...,N^u,
\end{equation}
\begin{equation}\label{condition2u}
\overline{p}_{i+1}^u-\underline{p}_{i+1}^u \sim
\log{n}\cdot\frac{\overline{p}_i^u-\underline{p}_i^u}{\kappa_i^u},
\quad\forall i = 1,...,N^u-1,
\end{equation}
\begin{equation}\label{condition3u} \tau_{i+1}^u\cdot \left(\frac{\overline{p}_i^u -
\underline{p}_i^u}{\kappa_i^u}\right)^2\cdot\sqrt{\log{n}} \sim
\tau_1^u, \quad\forall i = 1,...,N^u-1.
\end{equation}
Also we define
\begin{equation}\label{conditionforNu}
N^u = \min_{l}\left\{l:\left(\frac{\overline{p}_l^u -
\underline{p}_l^u}{\kappa_l^u}\right)^2\sqrt{\log{n}}<\tau_1^u\right\}.
\end{equation}

Next we state the set of relationships we want
$(\tau_i^c,\kappa_i^c)_{i=1}^{N^c}$ to satisfy:
\begin{equation}\label{condition1c}
\frac{\overline{p}_i^c-\underline{p}_i^c}{\kappa_i^c}\sim\sqrt{\frac{\kappa_i^c}{n\tau_i^c}},\quad\forall
i = 2,...,N^c,
\end{equation}
\begin{equation}\label{condition2c}
\overline{p}_{i+1}^c-\underline{p}_{i+1}^c \sim
\log{n}\cdot\frac{\overline{p}_i^c-\underline{p}_i^c}{\kappa_i^c},
\quad\forall i = 1,...,N^c-1,
\end{equation}
\begin{equation}\label{condition3c}
\tau_{i+1}^c \cdot \frac{\overline{p}_i^c -
\underline{p}_i^c}{\kappa_i^c}\cdot\sqrt{\log{n}} \sim
\tau_1^c,\quad\forall i = 1,...,N^c-1.
\end{equation}
Also we define
\begin{equation}\label{conditionforNc}
N^c = \min_{l}\left\{l:\frac{\overline{p}_l^c -
\underline{p}_l^c}{\kappa_l^c}\sqrt{\log{n}}<\tau_1^c\right\}.
\end{equation}
To understand the above relationships, it is useful to examine the
source of revenue losses in this algorithm. First, there is an {\it
exploration loss} in each period - the prices tested are not
optimal, resulting in suboptimal revenue rate or suboptimal
inventory consumption rate. The magnitude of such losses in each
period is roughly the deviation of the revenue rate (or the
inventory consumption rate) multiplied by the time length of the
period. Second, there is a {\it deterministic loss} due to the
limited learning capacity - we only test a grid of prices in each
period, and may never use the exact optimal price. Third, since the
demand follows a stochastic process, the observed demand rate may
deviate from the true underlying demand rate, resulting in a {\it
stochastic loss}. Note that these three losses also exist in the
learning algorithm proposed in \cite{besbes}. However, in dynamic
learning, the loss in one period does not simply appear once, it may
have impact on all the future periods. The design of our algorithm
tries to balance these losses in each step to achieve the maximum
efficiency of learning. With these in mind, we explain the meaning
of each equation above in the following:

\begin{itemize}
\item The first relationship (\ref{condition1u}) ((\ref{condition1c}), resp.) balances the deterministic loss induced by only
considering the grid points (the grid granularity is
$\frac{\overline{p}_i^u-\underline{p}_i^u}{\kappa_i^u}$
$(\frac{\overline{p}_i^c-\underline{p}_i^c}{\kappa_i^c}$, resp.))
and the stochastic loss induced in the learning period which will be
shown to be $\sqrt{\frac{\kappa_i^u}{n\tau_i^u}}$
($\sqrt{\frac{\kappa_i^c}{n\tau_i^c}}$, resp.). Due to the
relationship in (\ref{localpu}) and (\ref{localpc}), the loss is
quadratic in the price granularity in Step 2, and linear in Step 3.

\item The second relationship (\ref{condition2u}) ((\ref{condition2c}), resp.) is used to make sure
that with high probability, the price intervals $I_i^u$ ($I_i^c$,
resp.) contain the optimal price $p^D$. This has to be guaranteed,
otherwise a constant loss will be incurred in all periods
afterwards.

\item The third relationship (\ref{condition3u}) ((\ref{condition3c}), resp.) is used to bound
the exploration loss for each learning period. This is done by
considering the multiplication of the revenue rate deviation (also
demand rate deviation) and the length of the learning period, which
in our case can be upper bounded by
$\tau_{i+1}^u\sqrt{\log{n}}\cdot\left({\frac{\overline{p}_i^u -
\underline{p}_i^u}{\kappa_i^u}}\right)^2$
($\tau_{i+1}^c\sqrt{\log{n}}\cdot{\frac{\overline{p}_i^c -
\underline{p}_i^c}{\kappa_i^c}}$, resp.). We want this loss to be of
the same order for each learning period (and all equal to the loss
in the first learning period, which is $\tau_1$) to achieve the
maximum efficiency of learning.

\item Formula (\ref{conditionforNu}) ((\ref{conditionforNc}), resp.)
determines when the price we obtain is close enough to optimal such
that we can apply this price in the remaining selling season. We
show that $\sqrt{\log{n}}\cdot\left({\frac{\overline{p}_l^u -
\underline{p}_l^u}{\kappa_l^u}}\right)^2$
($\sqrt{\log{n}}\cdot{\frac{\overline{p}_l^c -
\underline{p}_l^c}{\kappa_l^c}}$, resp.) is an upper bound of the
revenue rate and demand rate deviation of price $\hat{p}_l$. When
this is less than $\tau_1$, we can simply apply $\hat{p}_l$ and the
loss will not exceed the loss of the first learning period.
\end{itemize}

\noindent Now we solve the relations
(\ref{condition1u})-(\ref{condition3u}) and obtain a set of
parameters that satisfy them:
\begin{eqnarray}\label{accuratetauu}
\tau_1^u = n^{-\frac{1}{2}}\cdot{(\log{n})}^{3.5}\mbox{  and  }
\tau_i^u =
n^{-\frac{1}{2}\cdot(\frac{3}{5})^{i-1}}\cdot{(\log{n})}^{5},\quad\forall
i = 2,...,N^u,
\end{eqnarray}
\begin{eqnarray}\label{accuratekappau}
\kappa_i^u =
n^{\frac{1}{10}\cdot(\frac{3}{5})^{i-1}}\cdot\log{n},\quad\forall i
= 1,2,...,N^u.
\end{eqnarray}
As a by-product, we have
\begin{eqnarray}\label{differencepu}
\overline{p}_i^u - \underline{p}_i^u =
n^{-\frac{1}{4}(1-(\frac{3}{5})^{i-1}))},\quad\forall i =
1,2,...,N^u.
\end{eqnarray}
Similarly, we solve the relations
(\ref{condition1c})-(\ref{condition3c}) and obtain a set of
parameters that satisfy them:
\begin{eqnarray}\label{accuratetauc}
\tau_1^c = n^{-\frac{1}{2}}\cdot{(\log{n})}^{2.5} \mbox{  and  }
\tau_i^c =
n^{-\frac{1}{2}\cdot(\frac{2}{3})^{i-1}}\cdot{(\log{n})}^{3},\quad\forall
i = 2,...,N^c,
\end{eqnarray}
\begin{eqnarray}\label{accuratekappac}
\kappa_i^c =
n^{\frac{1}{6}\cdot(\frac{2}{3})^{i-1}}\cdot\log{n},\quad\forall i =
1,2,...,N^c.
\end{eqnarray}
{\noindent and}
\begin{eqnarray}\label{differencepc}
\overline{p}_i^c - \underline{p}_i^c =
n^{-\frac{1}{2}(1-(\frac{2}{3})^{i-1})},\quad\forall i = 1,...,N^c.
\end{eqnarray}
Note that by (\ref{differencepu}) and (\ref{differencepc}), the
price intervals defined in our algorithm indeed shrink in each
iteration.

\section{Outlines of the Proof of Theorem \ref{th:maintheorem}}
\label{section:proof}

In this section, we provide an outline of the proof of Theorem
\ref{th:maintheorem}. We leave most of the technical details in the
Appendix, only the major steps are presented.

We first show that our algorithm will stop within a finite number of
iterations. We have the following lemma:

\begin{lemma}\label{lemma:boundedstep}
$N^u$ and $N^c$ defined in (\ref{conditionforNu}) satisfy: $N^u \le
3\log{n}$ and $N^c\le3\log{n}$, for $n\ge 3$.
\end{lemma}

{\noindent\bf Proof. } See Appendix \ref{appendix:boundedstep}.
$\Box$

Lemma \ref{lemma:boundedstep} provides an upper bound of the number
of iterations of our algorithm. In our analysis, we frequently need
to take a union bound over the number of iterations, and Lemma
\ref{lemma:boundedstep} will be used. For brevity, the condition
that $n\ge 3$ will be omitted in the future discussion.

In our algorithm, it is important to make sure that the
deterministic optimal price $p^D$ is always contained in our price
interval with high probability. This is because once our price
interval does not contain the deterministic optimal price, a
constant loss will be incurred for all periods afterwards, and the
algorithm will not be asymptotic optimal. The next lemma ensures the
inclusion of $p^D$ in all steps with high probability.

\begin{lemma}\label{lemma:containsprice}
Assume $p^D$ is the optimal price for the deterministic problem and
Assumption 1 hold for $\Gamma = \Gamma(M,K,m_L,m_U)$. Define $A_0$
to be the event satisfying the following conditions:
\begin{enumerate}
\item If we never enter Step 3, then
$p^D\in I_i^u$ for all $i=1,2,...,N^u$;
\item If Step 2 stops at $i_0$ and the algorithm enters Step 3, then $p^c\ge p^u$, $p^D\in I_i^u$ for
all $i=1,2,...,i_0$ and $p^D\in I_j^c$ for all $j = 1,2,...,N^c$.
\end{enumerate}
Then $P(A_0) = 1 - O\left(\frac{1}{n}\right)$.
\end{lemma}

{\noindent\bf Proof.} Here we present a sketch proof to show that
condition 1 holds with probability $1 - O\left(\frac{1}{n}\right)$.
The complete proof of this lemma is given in Appendix
\ref{appendix:proofofcontinsprice}.

We first show that if $p^D\in I_i^u$, then with probability
$1-O\left(\frac{1}{n^2}\right)$, $p^D\in I_{i+1}^u$. Define
\begin{eqnarray*}
u_n^i = 2\log{n}\cdot\max\left\{
\left(\frac{\overline{p}_i^u-\underline{p}_i^u}{\kappa_i^u}\right)^2,\sqrt{\frac{\kappa_i^u}{n\tau_i^u}}\right\}.
\end{eqnarray*}
Denote the unconstrained and constrained optimal solutions on the
current interval to be $p_i^u$ and $p_i^c$. We can show (the details
are in Appendix \ref{appendix:proofofcontinsprice}) that with
probability $1-O\left(\frac{1}{n^2}\right)$, $|\hat{p}_i^u -p_i^u| <
C\sqrt{u_n^i}$ and $|\hat{p}_i^c -p_i^c| < C\sqrt{u_n^i}$ (in our
analysis, for simplicity, we use $C$ to denote a generic constant
which only depends on the coefficient in $\Gamma$, $x$ and $T$).
Therefore, with probability $1-O\left(\frac{1}{n^2}\right)$,
$|\hat{p}_i-p^D| < C\sqrt{u_n^i}$. On the other hand, the next price
interval is centered near $\hat{p}_i$ with length of order
$\sqrt{\log{n}}$ greater than $\sqrt{u_n^i}$. Therefore, with
probability $1-O\left(\frac{1}{n^2}\right)$, $p^D\in I_{i+1}^u$.
Then we take a union bound over all $i$'s (at most $O(\log{n})$ of
them) and condition 1 holds with probability
$1-O\left(\frac{1}{n}\right)$. $\Box$

A corollary of Lemma \ref{lemma:containsprice} is that if $p^u \ge
p^c$, then with probability $1-O\left(\frac{1}{n}\right)$, our
algorithm will not enter Step 3. When $p^c > p^u$, however, it is
also possible that our algorithm will not enter Step 3, but we show
in that case, $p^u$ must be very close to $p^c$ so that the revenue
collected is still near-optimal.

Now we have proved that with high probability, $p^D$ will always be
in our price interval. We next analyze the revenue collected by this
algorithm and prove our main theorem.

We condition our analysis on the time the algorithm enters Step 3.
Define the following events:
\begin{eqnarray*}
B_1 & = & \{i_0=1\}\\
B_2 & = & \{i_0=2\}\\
&\vdots& \\
B_{N^u} & = & \{i_0=N^u\}\\
B_{N^u+1}&=&\{\mbox{The algorithm never enters Step 3}\}.
\end{eqnarray*}

Define $Y_{ij}^u$ to be a Poisson random variable with parameter
$n\lambda(p_{i,j}^u)\Delta_i^u$; $Y_{ij}^c$ to be a Poisson random
variable with parameter $n\lambda(p_{i,j}^c)\Delta_i^c$; $\hat{Y}^u$
to be a Poisson random variable with parameter
$n\lambda(\tilde{p})(1-t_{N^u}^u)$ and $\hat{Y}_i^c$ to be a Poisson
random variable with parameter
$n\lambda(\tilde{q})(1-t_{N^c}^c-t_i^u)$. Define events
\begin{eqnarray*}
A_1 &  = &\left\{\omega\in A_0:
\sum_{i=1}^{N^u}\sum_{j=1}^{\kappa_i^u} Y_{ij}^u < nx\right\},\\ A_2
& = & \left\{\omega\in A_0: \sum_{i=1}^{N^u}\sum_{j=1}^{\kappa_i^u}
Y_{ij}^u + \sum_{i=1}^{N^c} \sum_{j=1}^{\kappa_i^c} Y_{ij}^c <
nx\right\},
\end{eqnarray*}
where $A_0$ is the event defined in Lemma 2. The expected revenue
collected by DPA can be therefore bounded by
\begin{eqnarray}
\label{firstboundonYc} J_n^\pi(x,T,\lambda) \ge
E\left[\sum_{i=1}^{N^u}\sum_{j=1}^{\kappa_i^u}p_{i,j}^uY_{ij}^uI(\cup_{l=i}^{N^u+1}B_l)I(A_1)\right]
+
E\left[\tilde{p}\min\left(\hat{Y}^u,(nx-\sum_{i,j}Y_{ij}^u)^+\right)I(B_{N^u+1})\right]\nonumber\\
+E\left[\sum_{i=1}^{N^c}\sum_{j=1}^{\kappa_i^c}p_{i,j}^cY_{ij}^cI(\cup_{i=1}^{N^u}B_i)I(A_2)\right]
+ \sum_{l=1}^{N^u} E\left[\tilde{q}\min\left(\hat{Y}_l^c,
(nx-\sum_{i=1}^{l}\sum_{j=1}^{\kappa_i^u}Y_{ij}^u-\sum_{i=1}^{N^c}\sum_{j=1}^{\kappa_i^c}Y_{ij}^c)^+\right)I(B_l)\right].
\end{eqnarray}
In (\ref{firstboundonYc}), the first and third terms are lower
bounds of the expected revenue collected in Step 2 and 3 of DPA,
respectively. The second and fourth terms are lower bounds of the
expected revenue collected in Step 4(a) and (b) respectively. In the
following, we further analyze each term in (\ref{firstboundonYc}).
We show that, the revenue collected in each term is ``close'' to the
revenue generated by the optimal deterministic price $p^D$ in the
same period. We first prove the following lemma for the first term:

\begin{lemma}\label{lemma:part1c}
\begin{eqnarray}\label{learningrevenuelemmac}
E\left[\sum_{i=1}^{N^u}\sum_{j=1}^{\kappa_i^u}p_{i,j}^uY_{ij}^uI(\cup_{l=i}^{N^u+1}B_l)I(A_1)\right]\ge
\sum_{i=1}^{N^u}n\tau_i^u
p^D\lambda(p^D)P(\cup_{l=i}^{N^u+1}B_i)-Cn\tau_1^u\log{n}.
\end{eqnarray}
\end{lemma}

Lemma \ref{lemma:part1c} says that the revenue collected in Step 2
is close to the revenue that would have been collected if one uses
the deterministic optimal price $p^D$ for the same time period. The
proof of the lemma consists of two main steps. The first step is to
show that one can remove the indicator function $I(A_1)$ without
losing more than $Cn\tau_1^u\log{n}$, this is done by bounding the
probability of $A_1$ and some basic probability inequalities. The
second step is to show that the difference between the revenue
earned at $p_{i,j}^u$ and $p^D$ can be adequately bounded. For this,
we use the local quadratic property of the revenue function at
$p^D$. We give a sketch proof as follows.\\

{\noindent\bf Proof.} First, the left hand side of (\ref{learningrevenuelemmac}) can be written as
\begin{eqnarray}\label{firststep}
\sum_{i=1}^{N^u}\sum_{j=1}^{\kappa_i^u}\left(E\left[p_{i,j}^uY_{ij}^uI(\cup_{l=i}^{N^u+1}B_l)\right]
-
E\left[p_{i,j}^uY_{ij}^uI(\cup_{l=i}^{N^u+1}B_l)I(A_1^c)\right]\right).
\end{eqnarray}
For the first term, note that$Y_{ij}^u$ is independent of
$\cup_{l=1}^{i-1}B_l$, therefore independent of
$I(\cup_{l=i}^{N^u+1}B_l)$. Thus
\begin{eqnarray*}
E\left[p_{i,j}^uY_{ij}^uI(\cup_{l=i}^{N^u+1}B_l)\right] =
n\Delta_{i}^up_{i,j}^u\lambda(p_{i,j}^u) P(\cup_{l=i}^{N^u+1}B_l).
\end{eqnarray*}
Also, as shown in Lemma \ref{lemma:containsprice}, with probability
$1-O\left(\frac{1}{n}\right)$, $p^D$ is within the current price
interval, and by Assumption 1,
\begin{eqnarray*}
p_{i,j}^u\lambda(p_{i,j}^u)\ge p^D\lambda(p^D) - C(\bar{p}_i^u-\underline{p}_i^u)^2.
\end{eqnarray*}
Therefore, we have
\begin{eqnarray*}
\sum_{i=1}^{N^u}\sum_{j=1}^{\kappa_i^u}E\left[p_{i,j}^uY_{ij}^uI(\cup_{l=i}^{N^u+1}B_l)\right]\ge \sum_{i=1}^{N^u}n\tau_i^up^D\lambda(p^D)P(\cup_{l=i}^{N^u+1}B_l)-Cn\tau_1^u\log{n}
\end{eqnarray*}
where the last term is because of (\ref{differencepu}).

For the second term in (\ref{firststep}), we apply the Cauchy-Schwarz inequality:
\begin{eqnarray*}
E\left[Y_{ij}^uI(\cup_{l=i}^{N^u+1}B_l)I(A_1^c)\right]\le \sqrt{E\left[(Y_{ij}^u)^2I(\cup_{l=i}^{N^u+1}B_l)\right]}\cdot\sqrt{P(A_1^c)}.
\end{eqnarray*}
Since $Y_{ij}^u$ is a Poisson random variable, we have
\begin{eqnarray*}
\sqrt{E\left[(Y_{ij}^u)^2I(\cup_{l=i}^{N^u+1}B_l)\right]} \le \sqrt{E\left[(Y_{ij}^u)^2\right]}\le Cn\Delta_i^u\lambda(p_{i,j}^u).
\end{eqnarray*}
In Appendix \ref{appendix:proofofpart1c}, we show that $P(A_1^c)=O\left(\frac{1}{n}\right)$. Therefore,
\begin{eqnarray*}
\sum_{i=1}^{N^u}\sum_{j=1}^{\kappa_i^u}E\left[p_{i,j}^uY_{ij}^uI(\cup_{l=i}^{N^u+1}B_l)I(A_1^c)\right]\le \sum_{i=1}^{N^u}\sum_{j=1}^{\kappa_i^u}C\sqrt{n}\Delta_i^n \le Cn\tau_1^u\log n.
\end{eqnarray*}
And thus Lemma \ref{lemma:part1c} holds. $\Box$\\

%

Next we study the second term in (\ref{firstboundonYc}). We have the
following lemma:

\begin{lemma}\label{lemma:part2c}
\begin{eqnarray}\label{secondterm}
E\left[\tilde{p}\min(\hat{Y}^u,(nx-\sum_{i,j}Y_{ij}^u)^+)I(B_{N^u+1})\right]\ge
n(1-t^u_{N^u})p^D\lambda(p^D)\cdot P(B_{N^u+1})-Cn\tau_1^u\log{n}.
\end{eqnarray}
\end{lemma}

Lemma \ref{lemma:part2c} says that the revenue earned at step 4(a)
is close to the revenue that would have been collected if one uses
the deterministic optimal price $p^D$ for the same time period. To
show this lemma, the key is to show that the situations when the
sale terminates early are rare. That is, with high probability,
$\hat{Y}^u + \sum_{i,j}Y_{ij}^u \le nx$. This is done by analyzing
the tail probability of Poisson random variables. The detailed
analysis is given as follows.\\

{\noindent\bf Proof.} We start with the following transformation of
the left hand side of (\ref{secondterm})
\begin{eqnarray*}
&& E\left[\tilde{p}\min(\hat{Y}^u,(nx-\sum_{i,j}Y_{ij}^u)^+)I(B_{N^u+1})\right]\\
  &\ge & E\left[\tilde{p}\min(\hat{Y}^u,(nx-\sum_{i,j}Y_{ij}^u)^+)I(A_1)I(B_{N^u+1})\right]\\
&= & E\left[\tilde{p}(\hat{Y}^u-\max(\hat{Y}^u-(nx-\sum_{i,j}Y_{ij}^u)^+,0))I(A_1)I(B_{N^u+1})\right]\\
&\ge & E\left[\tilde{p}\hat{Y}^u I(A_1)I(B_{N^u+1})\right] -
E\left[\tilde{p}(\hat{Y}^u+\sum_{i,j}Y_{ij}^u-nx)^+I(B_{N^u+1})\right].
\end{eqnarray*}
We then show that
\begin{eqnarray}\label{relation1}
E\left[\tilde{p}\hat{Y^u} I(A_1)I(B_{N^u+1})\right]\ge
n(1-t_{N_u}^u)p^D\lambda(p^D)P(B_{N^u+1}) - Cn\tau_1^u
\end{eqnarray}
which means that the revenue collected in this period is close to
the revenue generated using $p^D$ for this period of time. Also we
show that
\begin{eqnarray}\label{relation2}
E\left[\tilde{p}(\hat{Y}^u+\sum_{i,j}Y_{ij}^u-nx)^+I(B_{N^u+1})\right]
\le Cn\tau_1^u\log{n}
\end{eqnarray}
which means that the loss due to over-consumption of the inventory
is small. The detailed proof of (\ref{relation1}) and
(\ref{relation2}) can be found in Appendix
\ref{appendix:proofofpart2c}. Combining (\ref{relation1}) and
(\ref{relation2}), we prove Lemma \ref{lemma:part2c}.

For the third term in (\ref{firstboundonYc}), we have

\begin{lemma}\label{lemma:part3c}
\begin{equation}\label{thirdterm}
\sum_{i=1}^{N^c}\sum_{j=1}^{\kappa_i^c}E[p_{i,j}^cY_{ij}^cI(\cup_{l=1}^{N^u}B_l)I(A_2)]\ge
\sum_{i=1}^{N^c}n\tau_i^cp^D\lambda(p^D)P(\cup_{l=1}^{N^u}B_l) -
Cn\tau_1^c\log{n}.
\end{equation}
\end{lemma}

And for the last term in (\ref{firstboundonYc}), we have

\begin{lemma}\label{lemma:part4c}
For each $l = 1,...,N^u$,
\begin{eqnarray}\label{forthterm}
E[\tilde{q}\min(\hat{Y}_l^c,
(nx-\sum_{i=1}^{l}\sum_{j=1}^{\kappa_i^u}Y_{ij}^u-\sum_{i=1}^{N^c}\sum_{j=1}^{\kappa_i^c}Y_{ij}^c)^+)I(B_l)]\ge
np^D\lambda(p^D)(1-t_l^u-t_{N^c}^c)P(B_l)-Cn\tau_1^u\log{n}.
\end{eqnarray}
\end{lemma}

The proof of Lemma \ref{lemma:part3c} and \ref{lemma:part4c} are
similar to the proof of Lemma \ref{lemma:part1c} and
\ref{lemma:part2c} and are given in Appendix
\ref{appendix:proofofpart3c} and \ref{appendix:proofofpart4c}.

Finally, we combine Lemma \ref{lemma:part1c}, \ref{lemma:part2c},
\ref{lemma:part3c} and \ref{lemma:part4c} by adding the right hand
side of (\ref{learningrevenuelemmac}), (\ref{secondterm}),
(\ref{thirdterm}) and (\ref{forthterm}). We have
\begin{eqnarray*}
J_n^\pi(x,T,\lambda) & \ge & \sum_{i=1}^{N^u}n\tau_i^u
p^D\lambda(p^D)P(\cup_{l=i}^{N^u+1}B_i) + p^D\lambda(p^D)\cdot
n(1-t^u_{N^u})P(B_{N^u+1})\\
&& + \sum_{i=1}^{N^c}n\tau_i^cp^D\lambda(p^D)P(\cup_{l=1}^{N^u}B_l)
+np^D\lambda(p^D)(1-t_l^u-t_{N^c}^c)P(B_l) -
Cn(\tau_1^u+\tau_c^u)\log{n}\\
& = & \sum_{l=1}^{N^u} P(B_l) \left(\sum_{i=1}^l n\tau_i^u
p^D\lambda(p^D) + \sum_{i=1}^{N^c}n\tau_i^cp^D\lambda(p^D) +
(1-t_l^u-t_{N^c}^c)np^D\lambda(p^D)\right) \\
& & + P(B_{N^u+1}) \left(\sum_{i=1}^{N^u} n\tau_i^up^D\lambda(p^D) +
n (1-t_{N^u}^u)p^D\lambda(p^D)\right) -
Cn(\tau_1^u+\tau_c^u)\log{n}\\
& = & np^D\lambda(p^D) - Cn(\tau_1^u+\tau_c^u)\log{n}.
\end{eqnarray*}
Therefore,
\begin{eqnarray*}
R_n^\pi(x,T;\lambda) \le 1 - \frac{np^D\lambda(p^D) -
Cn(\tau_1^u+\tau_c^u)\log{n}}{np^D\lambda(p^D)} \le C (\tau_1^u +
\tau_1^c) \cdot \log{n} = \frac{C(\log{n})^{4.5}}{\sqrt{n}},
\end{eqnarray*}
and we have proved Theorem 1.

\section{Lower Bound Example}
\label{section:lowerboundexample}

In Section \ref{section:mainresult} and \ref{section:proof}, we have
proposed a dynamic pricing algorithm and proved an upper bound of
$O^*(n^{-1/2})$ on its regret in Theorem \ref{th:maintheorem}. In
this section, we show that there exists a class of demand functions
satisfying our assumptions such that no pricing policy can achieve
an asymptotic regret less than $O^*(n^{-1/2})$. This lower bound
example provides a clear evidence that the upper bound is tight.
Therefore, our algorithm achieves nearly the ``best performance''
among all possible algorithms and closes the performance gap for
this problem.  Because our algorithm can be applied for both
parametric and non-parametric settings, it also closes the gap
between parametric and non-parametric learning for this problem.

\begin{theorem}\label{pro:lowerbound} {\bf(Lower bound example)} Let $\lambda(p;z)
= 1/2 + z - zp$ where $z$ is a parameter taking values in
$Z=[1/3,2/3]$ (we denote this demand function set by $\Lambda$).
Assume that $\underline{p} = 1/2$ and $\overline{p} = 3/2$. Also
assume that $x =2$ and $T=1$. Then we have
\begin{itemize}
\item This class of demand function satisfies Assumption 1. Furthermore, for any $z\in [1/3, 2/3]$, the optimal price $p^D$
always equals to $p^u$ and $p^D\in [7/8, 5/4]$.
\item For any admissible pricing policy $\pi$ and all $n\ge 1$,
\begin{eqnarray*}
 \sup_{z\in Z}R_n^\pi(x,T;z)\ge
\frac{1}{12(48)^2\sqrt{n}}.
\end{eqnarray*}
\end{itemize}
\end{theorem}
We first explain some intuitions behind this example. Note that all
the demand functions in $\Lambda$ cross at one common point, that
is, when $p = 1$, $\lambda(p;z) = 1/2$. Such a price is called an
``uninformative'' price in \cite{broder}. When there exists an
``uninformative'' price, experimenting at that price will not gain
information about the demand function. Therefore, in order to
``learn'' the demand function (i.e., the parameter $z$) and
determine the optimal price, one must at least perform some price
experiments at prices away from the uninformative price; on the
other hand, when the optimal price is indeed the uninformative
price, doing price experimentations away from the optimal price will
incur some revenue losses. This tension is the key reason for such a
lower bound for the loss and mathematically it is reflected in
statistical bounds on hypothesis testing. In the rest of this
section, we prove Theorem \ref{pro:lowerbound}. The proof resembles
the example discussed in \cite{broder} and \cite{besbes2}. However,
since our model is different from that in \cite{broder} and
\cite{besbes2}, they differ in several ways. We will discuss the
differences in the end of this section.

We first list some properties of the demand function set we defined
in Theorem \ref{pro:lowerbound}.

\begin{lemma}\label{lem:conditioncounterexample}
For the demand function defined in Theorem \ref{pro:lowerbound},
denote the optimal price $p^D$ under parameter $z$ to be $p^D(z)$.
We have:
\begin{enumerate}
\item $p^D(z) = (1+2z)/(4z)$
\item $p^D(z_0) = 1$ for $z_0 = 1/2$
\item $\lambda(p^D(z_0); z) = 1/2$ for all $z$
\item $-4/3\le r''(p;z)\le - 2/3$ for all $p,z$
\item $|p^D(z)-p^D(z_0)|\ge \frac{|z-z_0|}{4}$
\item $p^D(z)=p^u(z)$ for all $z$.
\end{enumerate}
\end{lemma}

For any policy $\pi$, and parameter $z$, let ${\cP}_z^\pi$ denote
the probability measure associated with the observations (the
process observed when using policy $\pi$) when the true demand
function is $\lambda(p;z)$ with $E_z^\pi$ being the corresponding
expectation operator. In order to quantify the tension mentioned
above, we need a notion of ``uncertainty'' about the unknown demand
parameter $z$. For this, we use the Kullback-Leibler (K-L)
divergence over two probability measures for a stochastic process.

Given $z$ and $z_0$, the K-L divergence between the two measures
$P_{z_0}^\pi$ and $P_{z}^\pi$ over time $0$ to $T$ is given by the
following (we refer to \cite{bremaud} for this definition):
\begin{eqnarray}\label{definitiondivergence}
{\cK} {({\cP}_{z_0}^\pi,{\cP}_z^\pi)} & = &
E_{z_0}^\pi\left[\int_{0}^{T=1} n
\lambda(p(s);z)\left[\frac{\lambda(p(s);z_0)}{\lambda(p(s);z)}\log{\frac{\lambda(p(s);z_0)}{\lambda(p(s);z)}}+
1- \frac{\lambda(p(s);z_0)}{\lambda(p(s);z)}\right]ds \right]\\%
& = & E_{z_0}^\pi\left[\int_0^1 \left\{
n\left(\frac{1}{2}+z_0-z_0p(s)\right)\left(-\log{\frac{\frac{1}{2}+z-zp(s)}{\frac{1}{2}+z_0-z_0p(s)}}-1\right)
+ n \left(\frac{1}{2} + z-zp(s)\right)\right\}ds \right].\nonumber
\end{eqnarray}
Note that the K-L divergence is a measure of distinguishability
between probability measures: if two probability measures are close,
then they have a small K-L divergence and vice versa. In terms of
pricing policies, a pricing policy $\pi$ is more likely to
distinguish between the case when the parameter is $z$ and the case
when the parameter is $z_0$ if the quantity
${\cK}(P_{z_0}^\pi,P_{z}^\pi)$ is large.

Now we show the following lemma, which gives a lower bound of the
regret for any policy in terms of the K-L divergence; this means a
pricing policy that is better capable to distinguish different
parameters will also be more costly.

\begin{lemma}\label{lem:learningiscostly} For any $z\in Z$, and any policy $\pi
\in{\cP}$, we have
\begin{equation}\label{learningiscostly}
{\cK} ({\cP}_{z_0}^\pi,{\cP}_z^\pi)\le 24n (z_0 - z)^2 R_n^\pi(x, T;
z_0),\\
\end{equation}
where $z_0=1/2$ and $R_n^\pi(x,T;z_0)$ is the regret function
defined in (\ref{regret}) with $\lambda$ being $\lambda(p;z_0)$.
\end{lemma}

{\noindent\bf Proof.} The proof attempts to bound the final term in
(\ref{definitiondivergence}) and is given in Appendix
\ref{appendix:learningiscostly}. $\Box$

Now we have shown that in order to have a policy that is able to
distinguish between two different parameters, one has to give up
some portion of the revenue. In the following lemma, we show that on
the other hand, if a policy is not able to distinguish between two
close parameters, then it will also incur a loss:

\begin{lemma}\label{lem:notlearningiscostly}
Let $\pi$ be any pricing policy that sets prices in
$[\underline{p},\overline{p}]$ and $p_\infty$. Define $z_0 = 1/2$
and $z_1^n = z_0 + \frac{1}{4n^{1/4}}$ (note $z_1^n\in [1/3,2/3]$
for all $n\ge 2$). We have for any $n\ge 2$
\begin{equation}\label{notlearningiscostely} R_n^\pi(x,T;z_0) +
R_n^\pi(x,T;z_1^n) \ge
\frac{1}{3(48)^2\sqrt{n}}e^{-{\cK}({\cP}_{z_0}^\pi,
{\cP}_{z_1^n}^\pi)}.
\end{equation}
\end{lemma}
{\noindent\bf Proof.} The proof uses similar ideas as discussed in
\cite{besbes2} and \cite{broder}. Here we give a sketch of the
proof. We define two non-intersecting intervals around $p^D(z_0)$
and $p^D(z_1^n)$. We show that when the true parameter is $z_0$,
pricing using $p$ in the second interval will incur a certain loss
and the same order of loss will be incurred if we use $p$ in the
first interval when the true parameter is $z_1^n$. At each time, we
treat our policy $\pi$ as a hypothesis test engine, which maps the
historic data to two actions:
\begin{itemize}
\item Choose a price in the first interval
\item Choose a price outside the first interval
\end{itemize}
Then we can represent the revenue loss by the ``accumulated
probability'' of committing errors in those hypothesis tests. By the
theory of the hypothesis test, one can lower bound the probability
of the errors for any decision rule. Thus we obtain a lower bound of
revenue loss for any pricing policy. The complete proof is referred
to Appendix \ref{appendix:notlearningiscostly}. $\Box$

Now we combine Lemma \ref{lem:learningiscostly} and
\ref{lem:notlearningiscostly}. By picking $z$ in Lemma
\ref{lem:learningiscostly} to be $z_1^n$ and add
(\ref{learningiscostly}) and (\ref{notlearningiscostely}) together,
we have:
\begin{eqnarray*}
 2 \{R_n^{\pi}(x,T;z_0) + R_n^{\pi}(x,T;z_1^n)\} & \ge & \frac{3}{32\sqrt{n}}{\cK}({\cP}_{z_0}^\pi,{\cP}_{z_1^n}^\pi) +
\frac{1}{3(48)^2\sqrt{n}}e^{-{\cK}({\cP}_{z_0}^\pi,
{\cP}_{z_1^n}^\pi)}\\
& \ge &
\frac{1}{3(48)^2\sqrt{n}}({\cK}({\cP}_{z_0}^\pi,{\cP}_{z_1^n}^\pi)+e^{-{\cK}({\cP}_{z_0}^\pi,{\cP}_{z_1^n}^\pi)})\\
& \ge & \frac{1}{3(48)^2\sqrt{n}}
\end{eqnarray*}
The last inequality is because for any number $w>0$, $w + e^{-w} \ge
1$. Therefore, for any admissible pricing policy $\pi$ and any $n$,
$$\sup_{\lambda\in \Lambda} R^\pi_n(x,T;\lambda) \ge
\frac{1}{12(48)^2\sqrt{n}}$$ and Theorem \ref{pro:lowerbound} is
proved.\\

{\noindent\bf Remark.} Our proof is similar to the proof of the
corresponding worst case examples in \cite{besbes} and
\cite{broder}, but different in several ways. First, in
\cite{besbes}, they considered only a finite number of possible
prices (though their proof is for a high-dimensional case, for the
sake of comparison, here we compare our theorem with theirs in the
one dimensional case). In our case, a continuous interval of prices
is allowed. Therefore, the admissible policy in our case is much
larger. And the K-L divergence function is thus slightly more
sophisticated than the one used in their proof. In fact, the
structure of our proof more closely resembles the one in
\cite{broder} where they consider a worst-case example for a general
parametric choice model. However, in their model, the time is
discrete. Therefore, a discrete version of the K-L divergence is
used and the analysis is based on the sum of the errors of different
steps. Our analysis can be viewed as a continuous-time extension of
the proof in \cite{broder}.

\section{Numerical Results}
\label{section:numericalresults}

In this section, we perform numerical tests to examine the
performance of our dynamic pricing algorithm and compare it to other
existing algorithms. We first provide some suggestions on the
implementation of our algorithm.

\subsection{Implementation Suggestions}
\label{subsection:implementation_suggestions}

The way the parameters of our DPA are defined in Section
\ref{section:mainresult} is mainly for the ease of asymptotic
analysis. In practice, we find several modifications to the
algorithm that could improve its performance. In all our numerical
tests in this section, we adopt the following modifications.

\begin{itemize}
\item {\it Definitions of $\tau_i^u, \kappa_i^u, \tau_i^c$ and
$\kappa_i^c$.} The main concern in directly applying
(\ref{accuratetauu}) - (\ref{differencepc}) in small-sized problems
is that the $\log{n}$ factors would play an overly dominant role. In
our numerical experiments, we find that by properly modifying these
values, the performance could be significantly improved. In
particular, we find the following set of values of $\tau_i^u,
\kappa_i^u, \tau_i^c$ and $\kappa_i^c$ give consistently good
results in all our numerical tests (for $n$ ranges from $10^2$ to
$10^7$):
\begin{eqnarray*}
\tau_i^u & = & n^{-\frac{1}{2}\cdot\left(\frac{3}{5}\right)^{i-1}}, \quad\quad i=1,2,...,N,\\
\kappa_i^u & = & \lfloor
n^{\frac{1}{10}\cdot\left(\frac{3}{5}\right)^{i-1}}\cdot\sqrt{\log{n}}\rfloor,  \quad\quad i=1,2,...,N^u,\\
\tau_i^c & = & n^{-\frac{1}{2}\cdot\left(\frac{2}{3}\right)^{i-1}},
 \quad\quad i=1,2,...,N^c,\\
\kappa_i^c & =& \lfloor
n^{\frac{1}{6}\cdot\left(\frac{2}{3}\right)^{i-1}}\cdot\frac{\log{n}}{3}\rfloor,
 \quad\quad i=1,2,...,N^c.
\end{eqnarray*}

\item {\it The regime switching condition (\ref{distinguish})}:
We find that the condition (\ref{distinguish}) might be too strict
in most of our tests problem (especially after we have reduced the
values of $\kappa_i^u$'s). Instead, we find that using the simple
condition $\hat{p}_i^c > \hat{p}_i^u$ works very well.

\item {\it The shrinking mechanism}: In our numerical tests, we find
that the shrinking mechanism for the price intervals defined in
(\ref{nextlowerboundu})-(\ref{nextupperboundu}) and
(\ref{nextlowerboundc})-(\ref{nextupperboundc}) are somewhat slow.
In our experiments, we find the following shrinking strategy which
shrinks the price interval faster performs consistently well:
\begin{eqnarray*}
\underline{p}_{i+1}^u & = & \hat{p}_i -
\frac{\sqrt{\log{n}}}{2}\cdot\frac{\bar{p}_i^u-\underline{p}_i^u}{\kappa_i^u},\\
\bar{p}_{i+1}^u & = & \hat{p}_i +
\frac{\sqrt{\log{n}}}{2}\cdot\frac{\bar{p}_i^u-\underline{p}_i^u}{\kappa_i^u},\\
\underline{p}_{i+1}^c & = & \hat{q}_i -
\frac{\log{n}}{9}\cdot\frac{\bar{p}_i^c-\underline{p}_i^c}{\kappa_i^c},\\
\bar{p}_{i+1}^c & = & \hat{q}_i +
\frac{\log{n}}{9}\cdot\frac{\bar{p}_i^c-\underline{p}_i^c}{\kappa_i^c}.
\end{eqnarray*}

\item {\it The starting interval in Step 3}: In DPA, when we enter Step 3, we restart from the initial interval $[\underline{p},
\bar{p}]$. This is only for the convenience of the proof and is not
necessary. In practice, one should start with the interval
$[\underline{p}_{i_0}, \bar{p}_{i_0}]$. This will also improve the
performance of the algorithm.
\end{itemize}

\subsection{Sample Runs of the Algorithm}
\label{subsection:sample run}

In this subsection, we show two sample runs of our algorithm to help the readers
further understand its features. The underlying demand functions for
these two examples are $\lambda_1(p) = 30 - 3p$ and $\lambda_2(p) =
80e^{-0.5p}$. In both cases, we assume the initial inventory level
$x = 20$, the selling horizon $T=1$, and the initial price interval
$[\underline{p},\overline{p}] = [0.1, 10]$. We show the results for the case with
$n = 10^5$ in Table \ref{table:samplerun1}
and \ref{table:samplerun2}.

In Table \ref{table:samplerun1} and \ref{table:samplerun2}, the
second row corresponds to the current step in the algorithm while
the third row represents the amount of time spent in this time
period. The fourth and fifth rows show the current price interval
and the number of prices tested in this period. The sixth and
seventh rows correspond to the empirical optimal $p^c$ and $p^u$ in
this time period. Note that after the algorithm enters Step 3, we no
longer compute $\hat{p}^u$. The last row gives the cumulative
revenue at the end of the corresponding period. The deterministic
upper bounds of the expected revenue given by (\ref{deterministic})
for the two cases are $7.50\times10^6$ and $5.54\times10^6$,
respectively.

\begin{table}\centering
\begin{tabular}{|c|c|c|c|c|c|}
  \hline
  Iter $\#$ & 1 & 2 & 3 & 4 & 5 \\ \hline
  Step $\#$ & Step 2 & Step 2 & Step 2 & Step 2 & Step 4 \\
  $\tau_i$& 0.0032 & 0.0316 & 0.126 & 0.289 & 0.551 \\
  $[\underline{p}_i,\bar{p}_i]$ & $[0.1, 10]$ & $[3.83, 5.74]$ & $[4.49, 5.07]$ & $[4.94, 5.07]$ & Apply $5.01$ \\
  $\kappa_i$ & 10 & 6 & 5 & 4 & N/A \\
  $\hat{p}^c$ & 3.07 & 3.93 & 4.49 & 4.94 & N/A \\
  $\hat{p}^u$ & 5.05 & 4.83 & 5.07 & 5.01 & N/A \\
  Cum. Rev. & $1.43e4$ & $2.48e5$ & $1.19e6$ & $3.35e6$ & $7.49e6$\\
  \hline
\end{tabular}\caption{Sample run of DPA with $\lambda(p) =
30-3p$}\label{table:samplerun1}
\end{table}

\begin{table}\centering
\begin{tabular}{|c|c|c|c|c|c|c|c|}
  \hline
  Iter $\#$ & 1 & 2 & 3 & 4 & 5 & 6 & 7\\ \hline
  Step $\#$ & Step 2 & Step 3 & Step 3 & Step 3 & Step 3 & Step 3 & Step 4 \\
  $\tau_i$& 0.0032 & 0.0032 & 0.0215 & 0.0774 & 0.182 & 0.321 & 0.392 \\
  $[\underline{p}_i,\bar{p}_i]$ & $[0.1, 10]$ & $[1.52, 3.48]$ & $[2.43, 2.91]$ & $[2.65,2.81]$ & [2.73,2.79] & [2.75, 2.78] & Apply $2.77$ \\
  $\kappa_i$ & 10 & 12 & 8 & 6 & 5 & 4 & N/A \\
  $\hat{p}^c$ & 3.07 & 2.58 & 2.74 & 2.75 & 2.78 & 2.77 & N/A \\
  $\hat{p}^u$ & 2.08 & N/A & N/A & N/A & N/A & N/A & N/A \\
  Cum. Rev. & $8.89e3$ & $1.81e4$ & $1.38e5$ & $5.72e5$ & $1.59e6$ & $3.39e6$ & $5.49e6$\\
  \hline
\end{tabular}\caption{Sample run of DPA with $\lambda(p) = 80\exp(-0.5p)$}\label{table:samplerun2}
\end{table}

In the first example, $\lambda_1(p) = 30 - 3p$, one can compute that
$p^D = p^u = 5
> p^c = 3$. Our algorithm never enters Step 3. In the
second example, $\lambda_2(p) = 80e^{-0.5p}$, and $p^D = p^c = 2.77
> p^u = 2$. The switching condition in (\ref{distinguish}) is triggered
at the first iteration and the algorithm enters Step 3 from then on.
In fact, in nearly all of our experiments, our algorithm can
distinguish these two scenarios immediately. That is, if $p^c >
p^u$, the switching condition will be triggered in the first few
iterations; otherwise, the switching condition will never be
triggered. In both examples, our algorithm runs 4-5 iterations of
price learning and then enters the final step (Step 4) in which the
estimated optimal price is applied for the remaining period. Several
important observations can be made from these two examples:
\begin{enumerate}
\item The time spent in each iteration is increasing (for the second example, it is increasing after the algorithm enters Step 3). This observation is consistent with our definition of $\tau_i^u$ and $\tau_i^c$. It is also intuitive. At the beginning, we have a very loose price interval with some
prices far from optimal, therefore, we cannot afford to spend too
much time in these iterations. As we narrow down the price range for
later iterations, we can spend more time on price experimentation
without incurring too much revenue losses. This also allows us to
test more intensely in the narrowed price interval , and therefore
enables us to find the ``optimal price'' more accurately.

\item The number of prices tested in each iteration is decreasing (for the second example, it is decreasing after the algorithm enters Step 3).
There are two reasons for this. First, the length of the price
interval is decreasing, therefore, we have a smaller ranges of price
to test in each iteration. Second, as the price range shrinks
towards the ``optimal price'' $p^D$, we need to perform more
intensive testing to distinguish the optimal price. Therefore, we
want to allocate even more time (relative to the length of learning
period) for each price and thus reduce the number of prices tested.

\item In both examples, the learning periods account for about half of the
entire time horizon. In particular, our algorithm can quickly shrink
the price interval into less than $5\%$ of the optimal price within
1/10 of the total time horizon in both examples. All the remaining
learning and doing are performed using close-to-optimal prices. This
capacity of our algorithm is one main reason to guarantee good
performances.
\end{enumerate}
In summary, our algorithm first distinguishes whether $p^c$ or $p^u$
is optimal, then constructs a shrinking series of price intervals
around the optimal price. In particular, as time evolves, our
algorithm tests fewer prices with a longer time on each price. And
finally, the estimated optimal price is applied approximately for
the latter half of the time horizon.

\subsection{Performance Analysis and Comparisons}
\label{subsection:comparison}

\begin{figure}
\centering
\includegraphics[scale = 0.25]{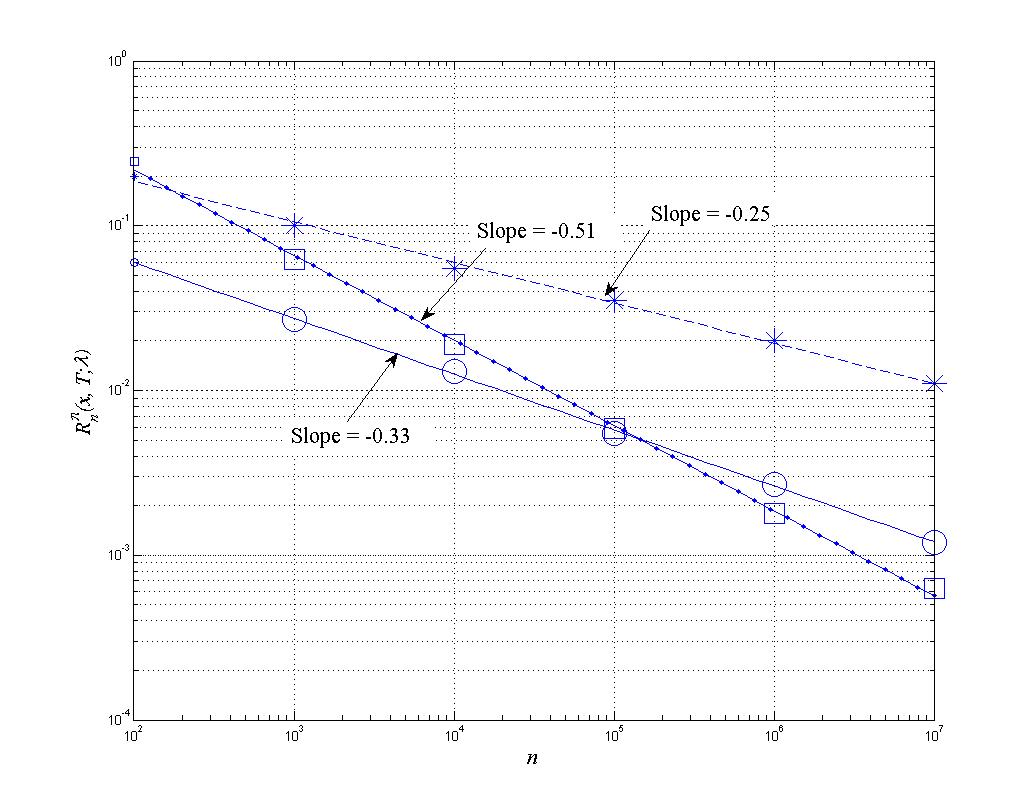}
\caption{Regret as a function of problem size for different
algorithms}\label{fig:comparison}
\end{figure}

We now perform extensive numerical performance tests of DPA and
compare against existing algorithms. Remember that the algorithms
are measured by the magnitude of the regret $R_n^{\pi}(x,
T;\lambda)$. And by Theorem \ref{th:maintheorem}, $R_n^{\pi}(x,
T;\lambda) =O^*(n^{-1/2})$. This implies that $\log{(R_n^{\pi}(x,
T;\lambda))}$ should be approximately a linear function of $\log{n}$
with slope close to $-1/2$. In Figure \ref{fig:comparison}, we show
a log-log plots of $R_n^{\pi}(x, T;\lambda)$ as a function of $n$
and compute a best linear trend line between $\log{(R_n^{\pi}(x,
T;\lambda))}$ and $\log{n}$. In Figure \ref{fig:comparison}, the
squares show the performance of our dynamic pricing algorithm. Stars
represent the regret of the non-parametric policy in \cite{besbes}.
Circles show the performance of the parametric policy in
\cite{besbes}. Lines represent the best line fit for the performance
of each policy. Each point in our numerical experiments is computed
by averaging regrets of 1000 repeated runs and the standard
deviation at each point is less than $0.1\%$ of the mean value.

In Figure \ref{fig:comparison}, we use the underlying demand
function $\lambda(p) = 30 - 3p$ with initial inventory $x = 20$, the
selling horizon $T=1$, and the initial price interval
$[\underline{p},\overline{p}] = [0.1, 10]$. We can see that the
asymptotic performance of our algorithm is indeed close to
$n^{-1/2}$, which numerically proves the validity of Theorem
\ref{th:maintheorem}. We also compare our algorithm to the
nonparametric and parametric policies introduced in \cite{besbes}.
One can observe that our algorithm outperforms the nonparametric
policies in \cite{besbes} for all $n$ greater than $100$. It also
outperforms the parametric policy in \cite{besbes} when $n\ge 10^5$,
despite the fact that the parametric policy knows the form of the
demand function while our algorithm has no information regarding the
demand. This verifies the efficiency of our algorithm.

\begin{figure}
\centering
\includegraphics[scale=0.4]{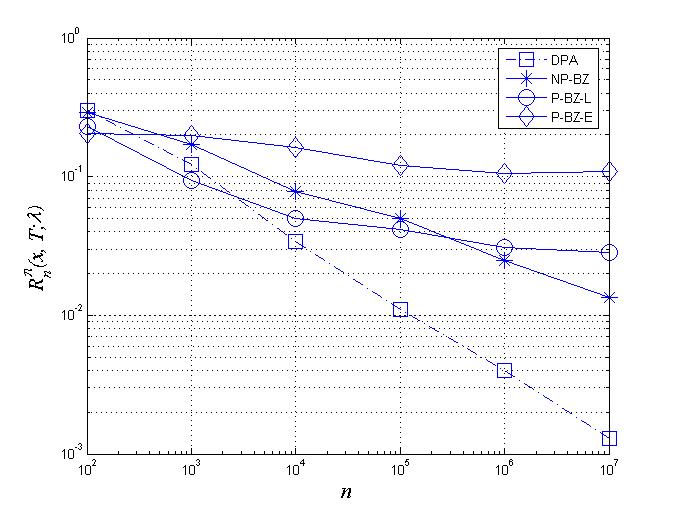}
\caption{Average performances of different
policies}\label{fig:comparison2}
\end{figure}

In the following, instead of specifying a certain underlying demand
function, we assume that the demand function is drawn
from a set of demand functions. We show that our algorithm behaves
robustly under demand function uncertainty. In particular, we
consider the set of demand functions consist of the two following
families:
\begin{itemize}
\item {\it Family 1}: $\lambda = a - bp$ with $a\in [20, 30]$, $b
\in [2, 10]$.
\item {\it Family 2}: $\lambda = a\exp(-bp)$ with $a\in [40, 80]$,
$b \in [1/3, 1]$.
\end{itemize}
We compare our dynamic pricing algorithm (DPA) with 1) Nonparametric
policy in \cite{besbes} (NP-BZ); 2) Parametric policy in
\cite{besbes} with assuming a linear demand function (P-BZ-L) and 3)
Parametric policy in \cite{besbes} with assuming an exponential
demand function (P-BZ-E). We still fix the initial inventory $x =
20$, the selling horizon $T=1$, and the initial price interval
$[\underline{p},\overline{p}] = [0.1, 10]$ and test for different
values of $n$. For each $n$, we run 1000 experiments, in each of
which the underlying demand function is chosen from family 1 and 2
each with probability $1/2$. The parameters $a$ and $b$ are chosen
uniformly at random within the ranges specified above. Note that the
set of demand functions include both the cases when $p^D = p^c >
p^u$ and $p^D=p^u> p^c$. The results are shown in Table
\ref{table:comparison} and Figure \ref{fig:comparison2} (the
standard errors of the numbers in Table \ref{table:comparison} are
less than $1\%$ of its value).

\begin{table}
\centering
\begin{tabular}{|c|cccc|cccc|}
  \hline
   & \multicolumn{4}{c|}{True Demand in Family 1} & \multicolumn{4}{c|}{True Demand in Family 2} \\ \hline
  $n$ & DPA & NP-BZ & P-BZ-L & P-BZ-E & DPA & NP-BZ & P-BZ-L & P-BZ-E
  \\ \hline
  $10^2$ & 0.3478 & 0.2739 & 0.0868 & 0.3374 & 0.253 & 0.3034 & 0.3737 & 0.0698 \\
  $10^3$ & 0.1601 & 0.1535 & 0.0705 & 0.3596 & 0.0845 & 0.1867 & 0.1158 & 0.0377  \\
  $10^4$ & 0.0383 & 0.0645 & 0.0327 & 0.3080 & 0.0298 & 0.0918 & 0.0668 & 0.0178  \\
  $10^5$ & 0.0127 & 0.0358 & 0.0111 & 0.2338 & 0.0101 & 0.0633 & 0.0725 & 0.0067  \\
  $10^6$ & 0.0041 & 0.0217 & 0.0047 & 0.2073 & 0.0038 & 0.0297 & 0.0574 & 0.0034  \\
  $10^7$ & 0.0013 & 0.0110 & 0.0021 & 0.2166 & 0.0013 & 0.0159 & 0.0548 & 0.0017 \\
  \hline
\end{tabular}\caption{Comparisons of the regrets for different
policies}\label{table:comparison}
\end{table}

In Table \ref{table:comparison} and Figure \ref{fig:comparison2}, we
can see that our dynamic pricing algorithm works robustly for all
underlying demand functions in the family specified above. The
log-regret still have a linear relationship with respect to $\log
{n}$ with a slope roughly being $-0.5$. The nonparametric policy in
\cite{besbes} also gives decreasing regret as $n$ grows. However,
the regret is much larger than ours. The performances of the
parametric policies are divided. They work well when the model is
well specified, that is, when the underlying demand function is
indeed as assumed. However, when the underlying demand function does
not belong to the assumed parametric family, the regrets typically
do not converge to $0$ as $n$ grows. This means that the risk of
model misspecification could be large. We also observe that the
regrets of the parametric policies are sensitive to the choice of
learning points. In the results presented, we choose learning points
$p_1=2$ and $p_2=3$, which yields the best performance among all the
combinations we tested, yet it performs much worse than the DPA as
$n$ grows large.

\section{Conclusion and Future Work}
\label{section:conclusion}

In this paper, we present a dynamic pricing algorithm for a class of
single-product revenue management problems. Our algorithm achieves
an asymptotic regret of $O^*(n^{-1/2})$ even if we have no prior
knowledge on the demand function except some regularity conditions.
By complementing with a worst-case bound, we show that the
convergence rate of our algorithm is among the fastest of all
possible algorithms in terms of asymptotic ¡°regret¡±. Our result
closes the performance gaps between parametric and non-parametric
learning and between a post-price mechanism and a customer-bidding
mechanism. Our dynamic pricing algorithm integrates learning and
doing in a concurrent procedure and provides important operational
insights to revenue management practitioners.

There are several future directions to explore. One possible
direction is to weaken the assumptions. Right now we assume the
revenue function has bounded second derivative. However, in
practice, the demand function might have a kink point (e.g., the
demand function is piecewise linear). In an extension of this work
\cite{wang_thesis}, we show that if we know the existence of such
kinks, we can modify our dynamic learning algorithm to accommodate
it, with the same asymptotic performance. Another direction is to
extend this result to high-dimensional problems. In high-dimensional
problems, the structure of the demand functions may be even more
complicated and the extension of dynamic learning is not
straightforward.

\section{Acknowledgement}
\label{section:acknowledgement} We would like to thank Professor
Peter W. Glynn for useful discussions. We also thank two
anonymous referees and the associate editor for the valuable comments
and suggestions which help us greatly improve our paper.

\bibliographystyle{plain}
\bibliography{paper}

\section{Appendix}
\label{section:appendix}

\subsection{A Lemma on the Deviation of Poisson Random Variables}
\label{appendix:poissonrv} In our proof, we will frequently use the
following lemma on the tail behavior of Poisson random variables:

\begin{lemma}\label{lemma:poissonrv}
Suppose that $\mu\in[0,M]$ and $r_n\ge n^\beta$ with $\beta>0$. If
$\epsilon_n = 2\eta^{1/2}M^{1/2}(\log{n})^{1/2}r_n^{-1/2}$, then for
all $n\ge 1$,
\begin{eqnarray*}
P(|N(\mu r_n) - \mu r_n| > r_n\epsilon_n)\le \frac{C}{n^\eta}.
\end{eqnarray*}
\end{lemma}

We refer to Lemma 2 in the online companion of \cite{besbes} for the
proof of this lemma.

\subsection{Proof of Lemma \ref{lemma:boundedstep}}
\label{appendix:boundedstep} By plugging (\ref{accuratekappau}) and
(\ref{differencepu}) into (\ref{conditionforNu}), we have $N^u =
\min_l\{l:n ^{\frac{1}{2}\cdot(\frac{3}{5})^l}<(\log{n})^{5}\}$. By
doing some algebra, we can get that
\begin{eqnarray}\label{boundonNu}
N^u\le  \log_{(5/3)}{\frac{\log{n}}{10\log\log{n}}}+ 1.
\end{eqnarray}
When $n\ge 3$,  the right hand side of (\ref{boundonNu}) is less than $\frac{\log{\log{n}}}{\log{(5/3)}} + 1.3$. It is easy to verify that this is less than $3\log{n}$ at $n=3$ and also has strictly smaller derivative than $3\log{n}$ for $n\ge 3$. Therefore, we must have $N^u\le 3\log{n}$ for $n\ge 3$.

Similarly, by plugging (\ref{accuratekappac}) and (\ref{differencepc}) into
(\ref{conditionforNc}), we have $N^c = \min_l\{l:n
^{\frac{1}{2}\cdot(\frac{2}{3})^l}<(\log{n})^{3}\}$. By doing some algebra, we can get that
\begin{eqnarray}\label{boundonNc}
N^c\le  \log_{(3/2)}{\frac{\log{n}}{6\log\log{n}}}+ 1.
\end{eqnarray}
When $n\ge 3$,  the right hand side of (\ref{boundonNc}) is less than $\frac{\log{\log{n}}}{\log{(3/2)}} + 2.5$. Again it is easy to verify that this is less than $3\log{n}$ at $n=3$ and also has strictly smaller derivative than $3\log{n}$ for $n\ge 3$. Therefore, we have $N^c\le 3\log{n}$ for $n\ge 3$.

\subsection{Proof of Lemma \ref{lemma:containsprice}}
\label{appendix:proofofcontinsprice}

{\bf Part 1:} We first prove that condition 1 holds with probability
$1-O\left(\frac{1}{n}\right)$. To this end, we show that if $p^D\in
I_i^u$, then with probability $1-O\left(\frac{1}{n^2}\right)$,
$p^D\in I_{i+1}^u$. We assume $p^D\in I_i^u$. Now consider the next
iteration. Define
$$u_n^i = 2\log{n}\cdot\max\left\{
\left(\frac{\overline{p}_i^u-\underline{p}_i^u}{\kappa_i^u}\right)^2,\sqrt{\frac{\kappa_i^u}{n\tau_i^u}}\right\}.$$
First, we establish a bound on the difference in the revenue rate
$r(\lambda(p^u_i))-r(\lambda(\hat{p}_i^u))$, where $p^u_i$ is the
revenue maximizing price on $I_i^u$ and $\hat{p}_i^u$ is the
empirical revenue maximizing price defined in our algorithm. We
assume $p_{i,j^*}^u$ is the nearest grid point to $p^u_i$ in this
iteration. We consider three cases:
\begin{itemize}
\item $p^u_i < p^u$: This is impossible since we know that $p^D
\ge p^u > p^u_i$ and by the induction assumption $p^D\in I_i^u$.
Therefore we must have $p^u\in I_i^u$, and by definition, $p^u$
achieves a larger revenue rate than $p^u_i$, which is contradictory
to the definition of $p_i^u$.
\item $p^u_i = p^u$: In this case, by the granularity of
the grid at iteration $i$, we have $|p_{i,j^*}^u - p^u|\le
\frac{\overline{p}_i^u-\underline{p}^u_i}{\kappa_i^u}$ and thus by
our assumption that $r''(\lambda)$ is bounded, we know that
$|r(\lambda(p^u)) - r(\lambda(p_{i,j^*}^u))|\le
m_LK^2\cdot\left(\frac{\overline{p}_i^u-\underline{p}_i^u}{\kappa_i^u}\right)^2$,
therefore we have:
\begin{align*}
& r(\lambda(p^u))-r(\lambda(\hat{p}^u_i)) \\
= & r(\lambda(p^u)) - r(\lambda(p_{i,j^*}^u)) +
p_{i,j}^u\lambda(p_{i,j^*}^u) -
p_{i,j^*}^u\hat{\lambda}(p_{i,j^*}^u) -
(\hat{p}^u_i\lambda(\hat{p}^u_i) -
\hat{p}^u_i\hat{\lambda}(\hat{p}^u_i)) +
p_{i,j^*}^u\hat{\lambda}(p_{i,j^*}^u) - \hat{p}^u_i \hat{\lambda}(\hat{p}^u_i)\\
\le &
m_LK^2\left(\frac{\overline{p}_i^u-\underline{p}_i^u}{\kappa_i^u}\right)^2
+ 2\max_{1\le j\le \kappa_i^u}|p_{i,j}^u\lambda(p_{i,j}^u) -
p_{i,j}^u\hat{\lambda}(p_{i,j}^u)|.
\end{align*}
where $\hat{\lambda}$ is the observed demand rate and the last
inequality is due to the definition of $\hat{p}^u_i$ and that
$\hat{p}^u_i$ is among one of the $p_{i,j}^u$.

By Lemma \ref{lemma:poissonrv} in Appendix \ref{appendix:poissonrv},
we have
\begin{eqnarray*}
P\left(|\hat{\lambda}(p_{i,j}^u)-\lambda(p_{i,j}^u)| >
C\sqrt{\log{n}}\cdot\sqrt{\frac{\kappa_i^u}{n\tau_i^u}}\right)\le
\frac{1}{n^2},
\end{eqnarray*}
with some suitable constant $C$. Therefore, with probability
$1-O(\frac{1}{n^2})$, $r(\lambda(p^u)) - r(\lambda(\hat{p}^u_i))\le
Cu_n^i$. However, by our assumption that $r''(\lambda)\le -m_U$ and
that $\gamma(\lambda)$ is Lipschitz continuous, with probability
$1-O\left(\frac{1}{n^2}\right)$, $|p^u - \hat{p}^u_i|\le
C\sqrt{u_n^i}$.

Now we consider the distance between $\hat{p}_i^c$ and $p_i^c$
(similar ideas can be found in Lemma 4 in the online companion of
\cite{besbes}). Assume $p_{i,j^*}$ is the nearest grid point to
$p_i^c$, we have:
\begin{eqnarray*}
|\lambda(\hat{p}_i^c)-x| & \le &
|\hat{\lambda}(\hat{p}_i^c)-x| +
|\hat{\lambda}(\hat{p}_i^c)-\lambda(\hat{p}_i^c)|\\
& \le & |\hat{\lambda}(p_{i,j^*}^u)-x| + |\hat{\lambda}(\hat{p}_i^c)-\lambda(\hat{p}_i^c)|\\
& \le & |\lambda(p_{i,j^*}^u)-x| + |\hat{\lambda}(p_{i,j^*}^u) -
\lambda(p_{i,j^*}^u)| + |\hat{\lambda}(\hat{p}_i^c)-\lambda(\hat{p}_i^c)|\\
& \le & |\lambda(p_i^c)-x| + |\lambda(p_i^c)-\lambda(p_{i,j^*}^u)| +
2\max_{1\le
j\le\kappa_i^c}|\hat{\lambda}(p_{i,j}^u)-\lambda(p_{i,j}^u)|.
\end{eqnarray*}
And by the definition of $p_i^c$, $\lambda(p_i^c)-x$ and
$\lambda(\hat{p}_i^c)-x$ must have the same sign, otherwise there
exists a point in between that achieves a smaller value of
$|\lambda(p)-x|$. Therefore,
\begin{eqnarray*}
|\lambda(p_i^c) - \lambda(\hat{p}_i^c)|\le
|\lambda(p_i^c)-\lambda(p_{i,j^*}^u)| + 2\max_{1\le
j\le\kappa_i^c}|\hat{\lambda}(p_{i,j}^u)-\lambda(p_{i,j}^u)|.
\end{eqnarray*}
By the Lipschitz continuity of $\lambda$, we have
\begin{eqnarray*}
|\lambda(p_i^c)-\lambda(p_{i,j^*}^u)|\le K\frac{\overline{p}_i^u
-\underline{p}_i^u}{\kappa_i^u}.
\end{eqnarray*}
Also by Lemma \ref{lemma:poissonrv} in Appendix
\ref{appendix:poissonrv}, we have with probability
$1-O\left(\frac{1}{n^2}\right)$,
\begin{eqnarray*}
\max_{1\le
j\le\kappa_i^c}|\hat{\lambda}(p_{i,j}^u)-\lambda(p_{i,j}^u)|\le
C\sqrt{\log{n}}\cdot \sqrt{\frac{\kappa_i^u}{n\tau_i^u}}.
\end{eqnarray*}

Therefore, with probability $1-O\left(\frac{1}{n^2}\right)$, $
|\lambda(p_i^c) - \lambda(\hat{p}_i^c)|\le C\sqrt{u_n^i}$ and by the
Lipschitz continuity of $\nu(\lambda)$, this implies that with
probability $1-O\left(\frac{1}{n^2}\right)$,
\begin{equation}\label{pcbound2}
|\hat{p}_i^c - p_i^c|\le C\sqrt{u_n^i}.
\end{equation}

Therefore, we have
\begin{eqnarray}\label{pbound}
P(|\hat{p}_{i} - p^D| > C\sqrt{u_n^i}) \le P(|\hat{p}_{i}^c - p_i^c|
> C\sqrt{u_n^i}) + P(|\hat{p}_{i}^u - p^u| > C \sqrt{u_n^i})\le
O\left(\frac{1}{n^2}\right).
\end{eqnarray}
Here we used the fact that:
\begin{eqnarray*}
|\max\{a,c\} - \max\{b,d\}| > u \quad \Rightarrow |a-b| > u \mbox{
or   } |c-d| > u.
\end{eqnarray*}
Note that (\ref{pbound}) is equivalent as saying that
\begin{equation*}
P\left(p^D\in[\hat{p}_i-C\sqrt{u_n^i},\hat{p}_i+C\sqrt{u_n^i}]\right)>1-O\left(\frac{1}{n^2}\right).
\end{equation*}
Now also note that the interval $I_{i+1}$ in our algorithm is chosen
to be
\begin{eqnarray*}
\left[\hat{p}_{i}-\frac{\log{n}}{3}\frac{\overline{p}_i^u -
\underline{p}_i^u}{\kappa_i^u}, \hat{p}_{i} +
\frac{2\log{n}}{3}\frac{\overline{p}_i^u -
\underline{p}_i^u}{\kappa_i^u}\right],
\end{eqnarray*}
which is of order
$\sqrt{\log{n}}$ greater than $\sqrt{u_n^i}$ (and according to the
way we defined $\kappa_i^u$ and $\tau_i^u$, the two terms in $u_n^i$
are of the same order). Therefore we know that with probability
$1-O(\frac{1}{n^2})$, $p^D\in I_{i+1}^u$.
\item $p^u < p^u_i$: In this case, $p^D=p^c$. With the same argument as above with only the $p^c$ part, we know that with
probability $1-O\left(\frac{1}{n^2}\right)$, $p^D\in I_{i+1}^u$
\end{itemize}

Also, as claimed in the previous lemma, the number of steps $N^u$ is
less than $3\log{n}$. Therefore, we can take a union bound over
$N^u$ steps, and claim that with probability
$1-O\left(\frac{1}{n}\right)$, $p^D\in I_i^u$, for all $i =
1,...,N^u$.

{\bf Part 2:} Using the same argument as in Part 1, we know that
with probability $1-O\left(\frac{1}{n}\right)$, $p^D\in I_i^u$ for
$i = 1,...,i_0$. Now at $i_0$, condition (\ref{distinguish}) is
triggered. We first claim that whenever (\ref{distinguish}) is
satisfied, with probability $1-O\left(\frac{1}{n}\right)$, $p^D=p^c
>p^u$.

By the argument in Part 1, we know that with probability
$1-O\left(\frac{1}{n}\right)$,
\begin{eqnarray*}|\hat{p}_{i_0}^c -
p_{i_0}^c|\le
\sqrt{\log{n}}\cdot\frac{\overline{p}_{i_0}^u-\underline{p}_{i_0}^u}{\kappa_{i_0}^u}\quad\mbox{
and }\quad |\hat{p}_{i_0}^u - p_{i_0}^u|\le
\sqrt{\log{n}}\cdot\frac{\overline{p}_{i_0}^u-\underline{p}_{i_0}^u}{\kappa_{i_0}^u}.
\end{eqnarray*}
Therefore, if
\begin{equation}\label{distinguish2}
\hat{p}_{i_0}^c > \hat{p}_{i_0}^u + 2\sqrt{
\log{n}}\cdot\frac{\overline{p}_{i_0}^u-\underline{p}_{i_0}^u}{\kappa_{i_0}^u},
\end{equation}
then with probability $1-O\left(\frac{1}{n}\right)$, $p_{i_0}^c >
p_{i_0}^u.$ holds. And when (\ref{distinguish2}) holds, $p_i^c$ must
not be the left end-point of $I_{i_0}^u$ and $p_{i_0}^u$ must not be
the right end-point of $I_{i_0}^u$, which means $p^u\le
p_{i_0}^u<p_{i_0}^c\le p^c= p^D.$

Now we consider the procedure in Step 3 of our algorithm and show
that with probability $1-O\left(\frac{1}{n}\right)$, $p^D = p^c \in
I_i^c$ for all $i = 1,2,...,N^c$.

We again prove that if $p^D\in I_i^c$, then with probability
$1-O\left(\frac{1}{n^2}\right)$, $p^D=p^c\in I_{i+1}^c$. We consider
$\hat{q}_i^c$ (which is the optimal empirical solution in Step 3).
Define in this case
$$ v_n^i = 2\sqrt{\log{n}}\cdot\max\left\{\frac{\overline{p}_i^c -
\underline{p}_i^c}{\kappa_i^c},
\sqrt{\frac{\kappa_i^c}{n\tau_i^c}}\right\}.$$ Using the same
discussion as in Part 1, we have with probability
$1-O\left(\frac{1}{n^2}\right)$,
$$|\hat{q}_i-p^c_i|=|\hat{p}_i^c - p^c| \le Cv_n^i.$$
However, in our algorithm, the next interval is defined as
$$I_{i+1}^c = \left[\hat{q}_i -
\frac{\log{n}}{2}\cdot\frac{\overline{p}_i^c-\underline{p}_i^c}{\kappa_i^c},
\hat{q}_i +
\frac{\log{n}}{2}\cdot\frac{\overline{p}_i^c-\underline{p}_i^c}{\kappa_i^c}
\right],$$ which is of order $\sqrt{\log{n}}$ larger than $v_n^i$.
Therefore with probability $1-O\left(\frac{1}{n^2}\right)$,
$p^D=p^c\in I_{i+1}^c$. Finally, taking a union bound over these
$N^c$ steps (no more than $\log{n}$) results in this lemma. $\Box$

\subsection{Proof of Lemma \ref{lemma:part1c}}
\label{appendix:proofofpart1c} We prove $P(A_1^c) = O(\frac{1}{n})$.
Note that we have already proved in Lemma 2 that $P(A_0) = 1 -
O(\frac{1}{n})$, therefore it suffices to show that
$P(\sum_{i,j}Y_{ij}^u > nx) = O\left(\frac{1}{n}\right)$. For this,
define $A_{ij}^u = \{\omega: Y_{ij}^u - EY_{ij}^u > EY_{ij}^u\}$. By
Lemma \ref{lemma:poissonrv}, we know that $P(A_{ij}^u) =
O\left(\frac{1}{n^3}\right)$. On the other hand,
\begin{eqnarray*}
\sum_{i,j} 2EY_{ij}^u \le  2M \sum_{i,j} n\Delta_i^u =
2Mn\sum_{i=1}^{N^u}\tau_i^u \le 2MnN^u\tau_{N^u}^u.
\end{eqnarray*}

By our definition of $\tau_i^u$, we know that $N^u\tau_{N^u}^u = o(1)$ (otherwise the algorithm would have stopped
at the previous iteration). Therefore, when $n$ is large enough,
$2\sum_{i,j}EY_{i,j}^u\le nx$. Thus
\begin{eqnarray*}
P(A_1^c) \le P(\cup_{i,j} A_{ij}^u) + P(A_0^c) = O\left(\frac{1}{n}\right).
\end{eqnarray*}%

\subsection{Proof of Lemma \ref{lemma:part2c}}
\label{appendix:proofofpart2c}

We start with the following transformation of the left hand side of
(\ref{secondterm}):
\begin{eqnarray*}
&& E[\tilde{p}\min(\hat{Y}^u,(nx-\sum_{i,j}Y_{ij}^u)^+)I(B_{N^u+1})]\nonumber\\
  &\ge & E[\tilde{p}\min(\hat{Y}^u,(nx-\sum_{i,j}Y_{ij}^u)^+)I(A_1)I(B_{N^u+1})]\nonumber\\
&= & E[\tilde{p}(\hat{Y}^u-\max(\hat{Y}^u-(nx-\sum_{i,j}Y_{ij}^u)^+,0))I(A_1)I(B_{N^u+1})]\nonumber\\
&\ge & E[\tilde{p}\hat{Y}^u I(A_1)I(B_{N^u+1})] -
E[\tilde{p}(\hat{Y}^u+\sum_{i,j}Y_{ij}^u-nx)^+I(B_{N^u+1})].
\end{eqnarray*}
For the first term, we have
\begin{eqnarray*}
E[\tilde{p}\hat{Y^u} I(A_1)I(B_{N^u+1})] = \tilde{p}E[\hat{Y}^uI(B_{N^u+1})] - \tilde{p}E[\hat{Y}^uI(B_{N^u+1})I(A_1^c)].
\end{eqnarray*}
However, $\tilde{p}E[\hat{Y}^uI(B_{N^u+1})] = n(1-t_{N^u}^u)\tilde{p}\lambda(\tilde{p})P(B_{N^u+1})$ since $\hat{Y}^u$ and $B_{N^u+1}$ are independent. Also by Lemma \ref{lemma:containsprice}, our assumption
on the bound of the second derivative of $r(\lambda)$,
(\ref{condition3u}) and (\ref{conditionforNu}), we have with probability $1-O\left(\frac{1}{n}\right)$
\begin{eqnarray*}
\tilde{p}\lambda(\tilde{p})\ge p^D\lambda(p^D) -
C(\overline{p}_{N^u+1}^u - \underline{p}_{N^u+1}^u)^2 \ge p^D\lambda(p^D) - C\tau_1^u.
\end{eqnarray*}
Therefore,
\begin{eqnarray*}
E[\tilde{p}\hat{Y^u}I(B_{N^u+1})]\ge
n(1-t_{N_u}^u)p^D\lambda(p^D)P(B_{N^u+1}) - Cn\tau_1^u.
\end{eqnarray*}
On the other hand, we have
\begin{eqnarray*}
\tilde{p}E[\hat{Y}^uI(B_{N^u+1})I(A_1^c)]\le \tilde{p}E[\hat{Y}^uI(A_1^c)]\le \tilde{p}\sqrt{E(\hat{Y}^u)^2}\sqrt{P(A_1^c)}\le C\sqrt{n}
\end{eqnarray*}
where the second inequality is because of Cauchy-Schwarz inequality and the last inequality is because $\hat{Y}^u$ is a Poisson random variable and $P(A_1^c)=O\left(\frac{1}{n}\right)$ as shown in Appendix
\ref{appendix:proofofpart1c}. Therefore,
\begin{eqnarray*}
E[\tilde{p}\hat{Y^u} I(A_1)I(B_{N^u+1})] \ge n(1-t_{N^u}^u)p^D\lambda(p^D)P(B_{N^u+1}) - Cn\tau_1^u.
\end{eqnarray*}

Next we consider
\begin{eqnarray*}
E[\tilde{p}(\hat{Y}^u+\sum_{i,j}Y_{ij}^u-nx)^+I(B_{N^u+1})].
\end{eqnarray*}
First we relax this to
\begin{eqnarray*}
\overline{p}E[(\hat{Y}^u+\sum_{i,j}Y_{ij}^u-nx)^+I(B_{N^u+1})].
\end{eqnarray*}

We have the following claims
\begin{claim}\label{lemma:deviationYu}
\begin{eqnarray*}
E(\hat{Y}^u+\sum_{i,j}Y_{ij}^u - E\hat{Y}^u - \sum_{i,j}EY_{ij}^u)^
+ \le Cn\tau_1^u \log{n},
\end{eqnarray*}
where $C$ is a constant only depending on the coefficients in
$\Gamma$, $x$ and $T$.
\end{claim}

\begin{claim}\label{lemma:expectedYu}
If $B_{N^u+1}$ occurs, then
\begin{eqnarray*}
\sum_{i=1}^{N^u}\sum_{j=1}^{\kappa_i^u}EY_{ij}^u + E\hat{Y}^u - nx
\le Cn\tau_1^u\log{n},
\end{eqnarray*}
where $C$ is a constant only depending on the coefficients in
$\Gamma$, $x$ and $T$.
\end{claim}

{\noindent\bf Proof of Claim \ref{lemma:deviationYu}.} We first show
that with probability $1-O(\frac{1}{n})$,
\begin{equation}\label{boundinventory}
\sum_{i,j}Y_{ij}^u+\hat{Y}^u-\sum_{i,j}EY_{ij}^u-E\hat{Y}^u\le
Cn\tau_1^u\log{n}.
\end{equation}
Then by Cauchy-Schwarz inequality,
\begin{eqnarray*}
E[(\hat{Y}^u+\sum_{i,j}Y_{ij}^u-E\hat{Y}^u-\sum_{i,j}EY_{ij}^u)^+ I(
\hat{Y}^u+&\sum_{i,j}Y_{ij}^u-E\hat{Y}^u-\sum_{i,j}EY_{ij}^u >
Cn\tau_1^u\log{n})]\nonumber\\
&\le O\left(\frac{1}{\sqrt{n}}\right)(E\hat{Y}^u +
\sum_{i,j}EY_{ij}^u)\le Cn\tau_1^u\log{n}
\end{eqnarray*}
where the second to last inequality is because $Y$'s are Poisson random variables. Therefore,
\begin{eqnarray*}
E(\hat{Y}^u+\sum_{i,j}Y_{ij}^u - E\hat{Y}^u - \sum_{i,j}EY_{ij}^u)^
+ \le  Cn\tau_1^u\log{n}.
\end{eqnarray*}

To show (\ref{boundinventory}), we apply Lemma
\ref{lemma:poissonrv}. For each given $i,j$, we have
$$P(Y_{ij}^u-EY_{ij}^u> 3M\sqrt{n\Delta_i^u\log{n}})\le \frac{C}{n^3}.$$
By taking a union bound over all $i$, $j$, we get
\begin{eqnarray*}
& P(\sum Y_{ij}^u - \sum EY_{ij}^u > 3M\sum_i
\kappa_i^u\sqrt{n\Delta_i^u\log{n}})\\
\le & \sum_{i=1}^{N^u}\sum_{j=1}^{\kappa_i^u} P(Y_{ij}^u-EY_{ij}^u >
3M\sqrt{n\Delta_i^u\log{n}})&\le O\left(\frac{1}{n}\right)
\end{eqnarray*}
where the last step is because $\kappa_i^u <n$ and $N^u \le 3\log{n}$.

On the other hand, by the definition of $\tau_i^u$ and $\kappa_i^u$,
we have
\begin{eqnarray*} 3M\sum_i \kappa_i^u\sqrt{n\Delta_i^u\log{n}} =
3M\sqrt{n\log{n}}\sum_{i=1}^{N^u}\sqrt{\kappa_i^u\tau_i^u}\le C N^u
n\tau_1^u \le Cn\tau_1^u\log{n},
\end{eqnarray*}
where the last inequality follows from the definition of $\tau_i^u$
and $\kappa_i^u$ in (\ref{accuratetauu}) and (\ref{accuratekappau}).
We then consider $\hat{Y}^u-E(\hat{Y}^u)$. Again we use Lemma
\ref{lemma:poissonrv}. We have
$$P(\hat{Y}^u-E(\hat{Y}^u)>2M\sqrt{n\log{n}})<\frac{C}{n^2},$$
since $\sqrt{n\log{n}} < n\tau_1^u$, the
claim holds.\quad\quad$\Box$\\

{\noindent\bf Proof of Claim \ref{lemma:expectedYu}.} By definition,
we have
$$EY_{ij}^u = n\lambda(p_{i,j}^u)\Delta_i^u,$$ where
$$ p_{i,j}^u = \underline{p}_i^u + (j-1)\frac{\overline{p}_i^u - \underline{p}_i^u}{\kappa_i^u} = \hat{p}_{i-1} -
\frac{\log{n}}{3}\cdot\frac{\overline{p}_{i-1}^u
-\underline{p}_{i-1}^u}{\kappa_{i-1}^u} + (j-1)
\frac{\overline{p}_i^u - \underline{p}_i^u}{\kappa_i^u}.$$ When
$B_{N^u+1}$ occurs, condition (\ref{distinguish}) doesn't hold,
i.e., $\hat{p}_{i-1} \ge \hat{p}_{i-1}^c - 2\sqrt{\log{n}}\cdot
\frac{\overline{p}_{i-1}^u-\underline{p}_{i-1}^u}{\kappa_{i-1}^u}$.
And as we showed in (\ref{pcbound2}), $\hat{p}_{i-1}^c \ge
p_{i-1}^c-\sqrt{\log{n}}\cdot
\frac{\overline{p}_{i-1}^u-\underline{p}_{i-1}^u}{\kappa_{i-1}^u}$.
Also since $p^u\ge p^c$, we must have $p_{i-1}^c\ge p^c$. Therefore
we have,
\begin{eqnarray*}
p_{i,j}^u & \ge & p^c + (j - 1)\frac{\overline{p}_i^u -
\underline{p}_i^u}{\kappa_i^u}
-\frac{\log{n}}{3}\cdot\frac{\overline{p}_{i-1}^u-\underline{p}_{i-1}^u}{\kappa_{i-1}^u}-3\sqrt{\log{n}}\cdot\frac{\overline{p}_{i-1}^u-\underline{p}_{i-1}^u}{\kappa_{i-1}^u}\nonumber\\
& \ge &  p^c  + (j - 1)\frac{\overline{p}_i^u -
\underline{p}_i^u}{\kappa_i^u} -
\frac{\log{n}}{2}\cdot\frac{\overline{p}_{i-1}^u-\underline{p}_{i-1}^u}{\kappa_{i-1}^u}\nonumber\\
\end{eqnarray*}
when $\sqrt{\log{n}}\ge 6$. Using the Taylor expansion for
$\lambda(p)$, we have that
\begin{eqnarray*}
\lambda(p_{i,j}^u) & \le &\lambda(p^c) + ((j-1)
\frac{\overline{p}_i^u - \underline{p}_i^u}{\kappa_i^u} -
\frac{\log{n}}{2}\cdot\frac{\overline{p}_{i-1}^u -
\underline{p}_{i-1}^u}{\kappa_{i-1}^u}) \lambda'(p^c) + \mbox{sup}
\{\lambda''(p^c)\}(\overline{p}_i^u - \underline{p}_i^u)^2\\ & \le &
\lambda(p^c) + C \cdot(\overline{p}_i^u-\underline{p}_i^u)^2.
\end{eqnarray*}
Therefore
\begin{eqnarray*}
\sum_{i,j} EY_{ij}^u\le n\lambda(p^c)t_{N^u}^u + Cn\sum_{i=1}^{N^u}
\tau_i^u(\overline{p}_i^u-\underline{p}_i^u)^2 \le
n\lambda(p^c)t_{N^u}^u+ Cn\tau_1^u\log{n},
\end{eqnarray*}
where the last equation follows from (\ref{condition2u}) and
(\ref{condition3u}).

Also we have $$E\hat{Y}^u = \lambda(\tilde{p})n(1-t_{N^u}^u),$$ and
with probability $1-O\left(\frac{1}{n}\right)$,
\begin{eqnarray*}
\tilde{p} =
\hat{p}_{N^u}+2\sqrt{\log{n}}\cdot\frac{\overline{p}_{N^u}^u-\underline{p}_{N^u}^u}{\kappa_i^{N^u}}\ge
\max(\hat{p}_{N^u}^u,\hat{p}_{N^u}^c) +
2\sqrt{\log{n}}\cdot\frac{\overline{p}_{N^u}^u-\underline{p}_{N^u}^u}{\kappa_i^{N^u}}
\ge p^D \ge p^c,
\end{eqnarray*}
where the first equation is due to the definition of $\tilde{p}$,
the second one is due to (\ref{pbound}), and the last one is by
Lemma \ref{lemma:containsprice}. Therefore,
$$ E\hat{Y}^u\le \lambda(p^c)n(1-t_{N^u}^u)$$
and thus
$$\sum_{i,j}E\hat{Y}_{ij} + E\hat{Y}^u \le nx+Cn\tau_1^u\log{n}.\quad\Box$$

With Claims \ref{lemma:deviationYu} and \ref{lemma:expectedYu},
Lemma \ref{lemma:part2c} follows.

\subsection{Proof of Lemma \ref{lemma:part3c}}
\label{appendix:proofofpart3c} First, the left hand side of
(\ref{thirdterm}) can be written as
\begin{eqnarray}\label{firststep2}
\sum_{i=1}^{N^c}\sum_{j=1}^{\kappa_i^c}\left(E\left[p_{i,j}^cY_{ij}^cI(\cup_{l=1}^{N^u}B_l)\right]
-
E\left[p_{i,j}^cY_{ij}^cI(\cup_{l=1}^{N^u}B_l)I(A_2^c)\right]\right).
\end{eqnarray}
For the first term, note that $Y_{ij}^c$ is independent with $\cup_{l=1}^{N^u}B_l$. Therefore,
\begin{eqnarray*}
E\left[p_{i,j}^cY_{ij}^cI(\cup_{l=1}^{N^u}B_l)\right] =
n\Delta_{i}^cp_{i,j}^c\lambda(p_{i,j}^c) P(\cup_{l=1}^{N^u}B_l).
\end{eqnarray*}
Also, by Lemma \ref{lemma:containsprice}, with probability $1-O\left(\frac{1}{n}\right)$,
\begin{eqnarray*}
p_{i,j}^c\lambda(p_{i,j}^c)\ge p^D\lambda(p^D) - C(\bar{p}_i^c-\underline{p}_i^c).
\end{eqnarray*}
Therefore, we have
\begin{eqnarray*}
\sum_{i=1}^{N^c}\sum_{j=1}^{\kappa_i^c}E\left[p_{i,j}^cY_{ij}^cI(\cup_{l=1}^{N^u}B_l)\right]\ge \sum_{i=1}^{N^c}n\tau_i^cp^D\lambda(p^D)P(\cup_{l=1}^{N^u}B_l)-Cn\tau_1^c\log{n}
\end{eqnarray*}
where the last term is because of (\ref{differencepc}).

For the second term in (\ref{firststep2}), we apply Cauchy-Schwarz inequality:
\begin{eqnarray*}
E\left[Y_{ij}^cI(\cup_{l=1}^{N^u}B_l)I(A_2^c)\right]\le \sqrt{E\left[(Y_{ij}^c)^2I(\cup_{l=1}^{N^u}B_l)\right]}\cdot\sqrt{P(A_2^c)}.
\end{eqnarray*}
Since $Y_{ij}^c$ is a Poisson random variable, we have
\begin{eqnarray*}
\sqrt{E\left[(Y_{ij}^c)^2I(\cup_{l=1}^{N^u}B_l)\right]} \le \sqrt{E\left[(Y_{ij}^c)^2\right]}\le Cn\Delta_i^c\lambda(p_{i,j}^c).
\end{eqnarray*}
The remaining task is to show that $P(A_2^c)=O\left(\frac{1}{n}\right)$. Once this holds, we have
\begin{eqnarray*}
\sum_{i=1}^{N^c}\sum_{j=1}^{\kappa_i^c}E\left[p_{i,j}^cY_{ij}^cI(\cup_{l=1}^{N^u}B_l)I(A_2^c)\right]\le
\sum_{i=1}^{N^c}\sum_{j=1}^{\kappa_c^u}C\sqrt{n}\Delta_i^c \le
Cn\tau_1^c\log n
\end{eqnarray*}
and Lemma \ref{lemma:part3c} will hold. To show $P(A_2^c)=O\left(\frac{1}{n}\right)$, it suffices to show that
$$P\left(\sum_{i=1}^{N^u}\sum_{j=1}^{\kappa_i^u}Y_{ij}^u + \sum_{i=1}^{N^c}\sum_{j=1}^{\kappa_i^c} Y{ij}^c > nx\right) = O\left(\frac{1}{n}\right).$$
For this, define $A_{ij}^u = \{\omega: Y_{ij}^u - EY_{ij}^u >
EY_{ij}^u\}$ and $A_{ij}^c = \{\omega: Y_{ij}^c - EY_{ij}^c >
EY_{ij}^c\}$. By Lemma \ref{lemma:poissonrv}, we know that
$P(A_{ij}^u) = O\left(\frac{1}{n^3}\right)$, $P(A_{ij}^c) = O\left(\frac{1}{n^3}\right)$. On the other
hand,
\begin{eqnarray*}
\sum_{i=1}^{N^u}\sum_{j=1}^{\kappa_i^u} EY_{ij}^u + \sum_{i=1}^{N^c}\sum_{j=1}^{\kappa_i^c} EY_{ij}^c \le  2Cn\left(\sum_{i=1}^{N^u}\sum_{j=1}^{\kappa_i^u}\Delta_i^u + \sum_{i=1}^{N^c}\sum_{j=1}^{\kappa_i^c} \Delta_i^c\right)  =
Cn\left(\sum_{i=1}^{N^u}\tau_i^u + \sum_{i=1}^{N^c}\tau_i^c\right).
\end{eqnarray*}
By our definition of $\tau_i^u$ and $\tau_i^c$ , we know that $\left(\sum_{i=1}^{N^u}\tau_i^u + \sum_{i=1}^{N^c}\tau_i^c\right)=o(1)$. Therefore, when $n$ is large enough,
$$\sum_{i=1}^{N^u}\sum_{j=1}^{\kappa_i^u} 2EY_{ij}^u + \sum_{i=1}^{N^c}\sum_{j=1}^{\kappa_i^c} 2EY_{ij}^c \le nx.$$
Thus $P(A_2) \le P(\cup_{i,j} A_{ij}^u) + P(\cup_{i,j} A_{ij}^c) +
P(A_0^c) = O(1/n)$. \hfill $\Box$

\subsection{Proof of Lemma \ref{lemma:part4c}}
\label{appendix:proofofpart4c} First, by doing a similar
transformation as in the proof of Lemma \ref{lemma:part2c}, we have
\begin{eqnarray*}
&&E\left[\tilde{q}\min\left(\hat{Y}_l^c,
(nx-\sum_{i=1}^{l}\sum_{j=1}^{\kappa_i^u}Y_{ij}^u-\sum_{i=1}^{N^c}\sum_{j=1}^{\kappa_i^c}Y_{ij}^c)^+\right)I(B_l)\right]\\
&\ge& E\left[\tilde{q}\hat{Y}_l^cI(B_l)I(A_2)\right] -
E\left[\tilde{q}\left(
\sum_{i=1}^l\sum_{j=1}^{\kappa_i^u}Y_{ij}^u + \sum_{i=1}^{N^c}\sum_{j=1}^{\kappa_i^c}Y_{ij}^c +\hat{Y}_l^c -
nx\right)^+I(B_l)\right].
\end{eqnarray*}
For the first term, we have
\begin{eqnarray*}
E[\tilde{q}\hat{Y}_l^cI(B_l)I(A_2)] \ge E[\tilde{q}\hat{Y}^cI(B_l)] - E[\tilde{q}\hat{Y}^cI(B_l)I(A_2^c)]
\end{eqnarray*}
However, $E[\tilde{q}\hat{Y}^cI(B_l)] = n(1-t_{N^c}^c-t_{l}^u)\tilde{q}\lambda(\tilde{q}) P(B_l)$ due to the independence between $\hat{Y}_l^c$ and $B_l$. Also by Lemma \ref{lemma:containsprice}, with probability $1-O\left(\frac{1}{n}\right)$,
\begin{eqnarray*}
\tilde{q}\lambda(\tilde{q})\ge p^D\lambda(p^D) - C(\bar{p}_{N^c+1}^c- \underline{p}_{N^c+1}^c)\ge p^D\lambda(p^D) - C\tau_1^c
\end{eqnarray*}
where the first inequality is because of the Lipschitz condition and the second one is because of the definition of $N^c$. Therefore,
\begin{eqnarray*}
E[\tilde{q}\hat{Y}_l^cI(B_l)]\ge n(1-t_{N^c}^c-t_{l}^u) p^D\lambda(p^D) - Cn\tau_1^c.
\end{eqnarray*}
On the other hand, we have
\begin{eqnarray*}
E[\tilde{q}\hat{Y}_l^cI(B_l)I(A_2^c)] \le \tilde{q}E[\hat{Y}_l^c I(A_2^c)] \le \tilde{q} \sqrt{E(\hat{Y}_l^c)^2} \sqrt{P(A_2^c)}.
\end{eqnarray*}
Since $\hat{Y}_l^c$ is a Poisson random variable, we have $\sqrt{E(\hat{Y}_l^c)^2}\le Cn$ and as have been proved in Appendix \ref{appendix:proofofpart3c}, $P(A_2^c) = O\left(\frac{1}{n}\right)$. Therefore,
\begin{eqnarray*}
E[\tilde{q}\hat{Y}_l^cI(B_l)I(A_2^c)]\le Cn\tau_1^c \quad\mbox{  and
}\quad E[\tilde{q}\hat{Y}_l^cI(B_l)I(A_2)] \ge
n(1-t_{N^c}^c-t_{l}^u) p^D\lambda(p^D) - Cn\tau_1^c.
\end{eqnarray*}
Next we consider
\begin{eqnarray*}
E\left[\tilde{q}\left(
\sum_{i=1}^l\sum_{j=1}^{\kappa_i^u}Y_{ij}^u +  + \sum_{i=1}^{N^c}\sum_{j=1}^{\kappa_i^c}Y_{ij}^c + \hat{Y}_l^c -
nx\right)^+I(B_l)\right].
\end{eqnarray*}
We first relax it to
$$\bar{p}E\left[\left(
\sum_{i=1}^l\sum_{j=1}^{\kappa_i^u}Y_{ij}^u +  + \sum_{i=1}^{N^c}\sum_{j=1}^{\kappa_i^c}Y_{ij}^c + \hat{Y}_l^c -
nx\right)^+I(B_l)\right]$$
By using the same arguments as Appendix \ref{appendix:proofofpart2c}, we know that
\begin{eqnarray*}
\bar{p}E\left[\left(
\sum_{i=1}^l\sum_{j=1}^{\kappa_i^u}Y_{ij}^u +  + \sum_{i=1}^{N^c}\sum_{j=1}^{\kappa_i^c}Y_{ij}^c + \hat{Y}_l^c -
nx\right)^+I(B_l)\right] \le Cn\tau_1^c\log{n}.
\end{eqnarray*}
Therefore, we have
\begin{eqnarray*}
E\left[\tilde{q}\min\left(\hat{Y}_l^c,
(nx-\sum_{i=1}^{l}\sum_{j=1}^{\kappa_i^u}Y_{ij}^u-\sum_{i=1}^{N^c}\sum_{j=1}^{\kappa_i^c}Y_{ij}^c)^+\right)I(B_l)\right] \ge n(1-t_{N^c}^c-t_{l}^u) p^D\lambda(p^D) - Cn\tau_1^c.
\end{eqnarray*}

\subsection{Proof of Lemma \ref{lem:learningiscostly}}
\label{appendix:learningiscostly} Consider the final term in
(\ref{definitiondivergence}). Note that we have the following simple
inequality:

$$\log{(x+1)}\ge x - \frac{x^2}{2(1-|x|)},\quad\forall x<1.$$
Therefore, we have $$- \log{x} - 1 \le - x +
\frac{(x-1)^2}{2(2-x)}\quad\forall\mbox{ } 0<x<2.$$ Apply this
relationship to the final term in (\ref{definitiondivergence}) and
note that for any $z\in [1/3,2/3]$ and $p(s)\in [1/2,3/2]$,
\begin{equation}\label{term1}
\frac{2}{3}\le \frac{\frac{1}{2} +
z-zp(s)}{\frac{1}{2}+z_0-z_0p(s)}\le 2,
\end{equation}
we have
\begin{eqnarray*}
{\cK}({\cP}_{z_0}^\pi,{\cP}_z^\pi)\le nE_{z_0}^\pi
\int_{0}^{1}\frac{1}{2(2-\frac{1/2+z_0-z_0p(s)}{1/2+z-zp(s)})}\frac{(z_0-z)^2(1-p(s))^2}{(1/2+z_0-z_0p(s))^2}ds\le
nE_{z_0}^\pi\int_0^1\frac{(z_0-z)^2(1-p(s))^2}{(1/2+z_0-z_0p(s))^2}ds.
\end{eqnarray*}
Also, for $z\in [1/3,2/3]$ and $p(s)\in[1/2,3/2]$, we have
\begin{eqnarray*}
\frac{1}{2}+z_0-z_0p(s)\ge \frac{1}{4}.
\end{eqnarray*}
Therefore, we have
$${\cK}({\cP}_{z_0}^\pi,{\cP}_z^\pi)\le16n(z_0-z)^2E_{z_0}^\pi\int_0^1 (1-p(s))^2ds.$$ However, under the case when the
parameter is $z_0$, we have $p^D = 1$ and
\begin{eqnarray*}
R_n^\pi(x,T;z_0) & = & 1- \frac{J_n^\pi(x,T;z_0)}{J_n^D(x,T;z_0)}
\ge \frac{E_{z_0}^\pi \int_0^1
(r(p^D)-r(p(s)))ds}{E_{z_0}^\pi\int_0^1 r(p^D)ds} \ge \frac{2}{3}
E_{z_0}^\pi \int_0^1 (1-p(s))^2 ds
\end{eqnarray*}
where the first inequality follows from the definition of $J^D$ and
that we relaxed the inventory constraint, and the second inequality
is because of the 4th condition in Lemma
\ref{lem:conditioncounterexample}. Therefore,
\begin{eqnarray*}
{\cK}({\cP}_{z_0}^\pi,{\cP}_z^\pi)\le 24 n(z_0 - z)^2 R_n^\pi(x, T;
z_0),
\end{eqnarray*}
and Lemma \ref{lem:learningiscostly} holds. $\Box$

\subsection{Proof of Lemma \ref{lem:notlearningiscostly}}
\label{appendix:notlearningiscostly} We define two intervals
$C_{z_0}$ and $C_{z_1^n}$ as follows:
$$C_{z_0} =
[p^D(z_0)-\frac{1}{48n^{1/4}},p^D(z_0)+\frac{1}{48n^{1/4}}]\mbox{
and } C_{z_1^n} = [p^D(z_1^n)-\frac{1}{48n^{1/4}}, p^D(z_1^n) +
\frac{1}{48n^{1/4}}].$$ Note that by the 5th property in Lemma
\ref{lem:conditioncounterexample}, we know that $C_{z_0}$ and
$C_{z_1^n}$ are disjoint.

By the 4th property in Lemma \ref{lem:conditioncounterexample}, we
have for any $z$,
\begin{equation}\label{instantenousregret}
r(p^D(z);z)-r(p;z)\ge \frac{2}{3}(p-p^D(z))^2.
\end{equation}
Also by the definition of the regret function, we have
\begin{eqnarray}\label{regret1}
R_n^\pi(x,T;z_0)&\ge&\frac{E_{z_0}^\pi\int_0^1\{r(p^D(z_0);z_0)-r(p(s);z_0)\}ds}{E_{z_0}^\pi
\int_0^1 r(p^D(z_0);z_0) ds}\nonumber\\
&\ge& \frac{4}{3(48)^2\sqrt{n}}E_{z_0}^\pi\int_0^1 I(p(s)\in
C_{z_1^n})
ds\nonumber\\
&= &  \frac{4}{3(48)^2\sqrt{n}}\int_0^1 {\cP}_{z_0}^{\pi(s)}
(p(s)\in C_{z_1^n}) ds,
\end{eqnarray}
where the first inequality is because we relaxed the inventory
constraint when using $\pi$, and the second inequality is because of
(\ref{instantenousregret}), the definition of $C_{z_1^n}$ and that
the denominator is 1/2. In the last equality, ${\cP}_{z_0}^{\pi(s)}$
is the probability measure under policy $\pi$ and up to time $s$
(with underlying demand function has parameter $z_0$). Similarly, we
have
\begin{eqnarray}\label{regret2} R_n^\pi(x,T;z_1^n)&\ge &
\frac{E_{z_1}^\pi\int_0^1\{r(p^D(z_1^n);z_1^n)-r(p(s);z_1^n)\}ds}{E_{z_1^n}^\pi
\int_0^1 r(p^D(z_1^n);z_1^n) ds}\nonumber\\
&\ge& \frac{2}{3(48)^2\sqrt{n}}E_{z_1^n}^\pi\int_0^1 I(p(s)\notin
C_{z_1^n})
ds \nonumber\\
& = & \frac{2}{3(48)^2\sqrt{n}}\int_0^1 {\cP}_{z_1^n}^{\pi(s)}
(p(s)\notin C_{z_1^n}) ds,
\end{eqnarray}
where in the second inequality, we use the fact that the denominator
is less than $1$, and in the last equality, ${\cP}_{z_1}^{\pi(s)}$
is defined similarly as the one in (\ref{regret1}).

Now consider any decision rule that maps historical demands up to
time $s$ into one of the following two sets:
$$H_1: p(s)\in C_{z_1^n}$$
$$H_2: p(s)\notin C_{z_1^n}.$$
By Theorem 2.2 in \cite{tsybakov}, we have the following bound on
the probability of error of any decision rule:
\begin{eqnarray*}
{\cP}_{z_0}^{\pi(s)}\{p(s)\in C_{z_1^n}\} +
{\cP}_{z_1^n}^{\pi(s)}\{p(s)\notin C_{z_1^n}\} \ge \frac{1}{2}
e^{-{\cK}({\cP}_{z_0}^{\pi(s)},{\cP}_{z_1^n}^{\pi(s)})}.
\end{eqnarray*}
However, by the definition of the K-L divergence
(\ref{definitiondivergence}), we know that
\begin{eqnarray*}
{\cK}({\cP}_{z_0}^{\pi},{\cP}_{z_1^n}^{\pi}) -
{\cK}({\cP}_{z_0}^{\pi(s)},{\cP}_{z_1^n}^{\pi(s)}) \ge E_{z_0}^\pi
\int_{s}^{1}\frac{1}{2(2-\frac{1/2+z_0-z_0p(s)}{1/2+z-zp(s)})}\frac{(z_0-z)^2(1-p(s))^2}{(1/2+z_0-z_0p(s))^2}ds\ge
0,
\end{eqnarray*} where the last inequality is because of (\ref{term1}). Therefore, we have
\begin{eqnarray*}
{\cP}_{z_0}^{\pi(s)}\{p(s)\in C_{z_1^n}\} +
{\cP}_{z_1^n}^{\pi(s)}\{p(s)\notin C_{z_1^n}\}\ge \frac{1}{2}
e^{-{\cK}({\cP}_{z_0}^{\pi},{\cP}_{z_1^n}^{\pi})}.
\end{eqnarray*}
Now we add (\ref{regret1}) and (\ref{regret2}) together. We have
\begin{eqnarray}\label{combinedregret}
R_n^\pi(x,T;z_0) + R_n^\pi(x,T;z_1^n) & \ge &
\frac{2}{3(48)^2\sqrt{n}}\int_0^1 \{{\cP}_{z_0}^{\pi(s)}(p(s)\in
C_{z_1^n})
+ {\cP}_{z_1^n}^{\pi(s)}(p(s)\notin C_{z_1^n})\}ds\nonumber\\
& \ge &
\frac{1}{3(48)^2\sqrt{n}}e^{-{\cK}({\cP}_{z_0}^{\pi},{\cP}_{z_1^n}^{\pi})}.\nonumber
\end{eqnarray}
Thus, Lemma \ref{lem:notlearningiscostly} holds. $\Box$

\end{document}